\documentclass[10pt,final,twocolumn,letterpaper]{article}
\usepackage[final]{cvpr}
\usepackage{url}
\usepackage{bm}
\usepackage{graphicx}
\usepackage{tabularx}       
\usepackage{comment}
\usepackage{makecell}
\usepackage{pifont}
\usepackage{booktabs}
\usepackage{multirow}

\usepackage[misc]{ifsym}
\usepackage{afterpage}
\usepackage{setspace}
\usepackage{breakcites}
\usepackage{seqsplit}

\usepackage{times}
\usepackage{epsfig}
\usepackage{graphicx}
\usepackage{amsmath}
\usepackage{amssymb}
\usepackage{float}
\usepackage{utfsym}
\usepackage[symbol]{footmisc}

\usepackage{hyphenat}
\usepackage{soul,color}
\usepackage{amsopn}

\definecolor{Red}{cmyk}{0,1,1,0}
\definecolor{Green}{cmyk}{1,0,1,0}
\definecolor{Cyan}{cmyk}{1,0,0,0}
\definecolor{Purple}{cmyk}{0.45,0.86,0,0}
\definecolor{Rosolic}{cmyk}{0.00,1.00,0.50,0}
\definecolor{Blue}{cmyk}{1.00,1.00,0.00,0}
\definecolor{BlueViolet}{cmyk}{0.86,0.91,0,0.04}
\definecolor{NavyBlue}{cmyk}{0.94,0.54,0,0}

\newcommand{\hidden}[1]{{\color{NavyBlue}}}
\newcommand{\crossmark}{\scalebox{0.75}{\usym{2613}}}
\definecolor{cvprblue}{rgb}{0.21,0.49,0.74}
\usepackage[pagebackref,breaklinks,colorlinks,citecolor=cvprblue]{hyperref}
\usepackage[capitalize]{cleveref}
\crefname{section}{Sec.}{Secs.}
\Crefname{section}{Section}{Sections}
\Crefname{table}{Table}{Tables}
\crefname{table}{Tab.}{Tabs.}

\begin{document}

%%%%%%%%% TITLE
\title{Paint3D: Paint Anything 3D with Lighting-Less Texture Diffusion Models}
% \title{Painter3D: Paint Anything 3D with Texture Diffusion Models}
% \title{PAY3D: Paint Anything 3D with Texture Diffusion Models}
% \title{TextureAnything: Control 3D Texture Generation via Text-to-Image Diffusion Models}

\author{
Xianfang Zeng$^{1}$\footnotemark[1]\quad\quad\quad\quad 
Xin Chen$^{1}$\footnotemark[1] \quad\quad\quad\quad
Zhongqi Qi$^{1}$\footnotemark[1] \quad\quad\quad\quad
Wen Liu$^{1}$  \quad\quad\quad\quad 
Zibo Zhao$^{1,3}$ \\ 
Zhibin Wang$^{1}$ \quad\quad\quad\quad 
BIN FU$^{1}$ \quad\quad\quad\quad 
Yong Liu$^{2}$ \quad\quad\quad\quad 
Gang Yu$^{1}$\footnotemark[2]\\ 
$^{1}$Tencent PCG \quad\quad\quad\quad $^{2}$Zhejiang University  \quad\quad\quad\quad $^{3}$ ShanghaiTech University \\
\tt \small \textbf{\href{https://github.com/OpenTexture/Paint3D}{https://github.com/OpenTexture/Paint3D}}
}

\makeatletter
\let\@oldmaketitle\@maketitle%
\renewcommand{\@maketitle}{\@oldmaketitle%
 \centering
    \includegraphics[width=\textwidth]{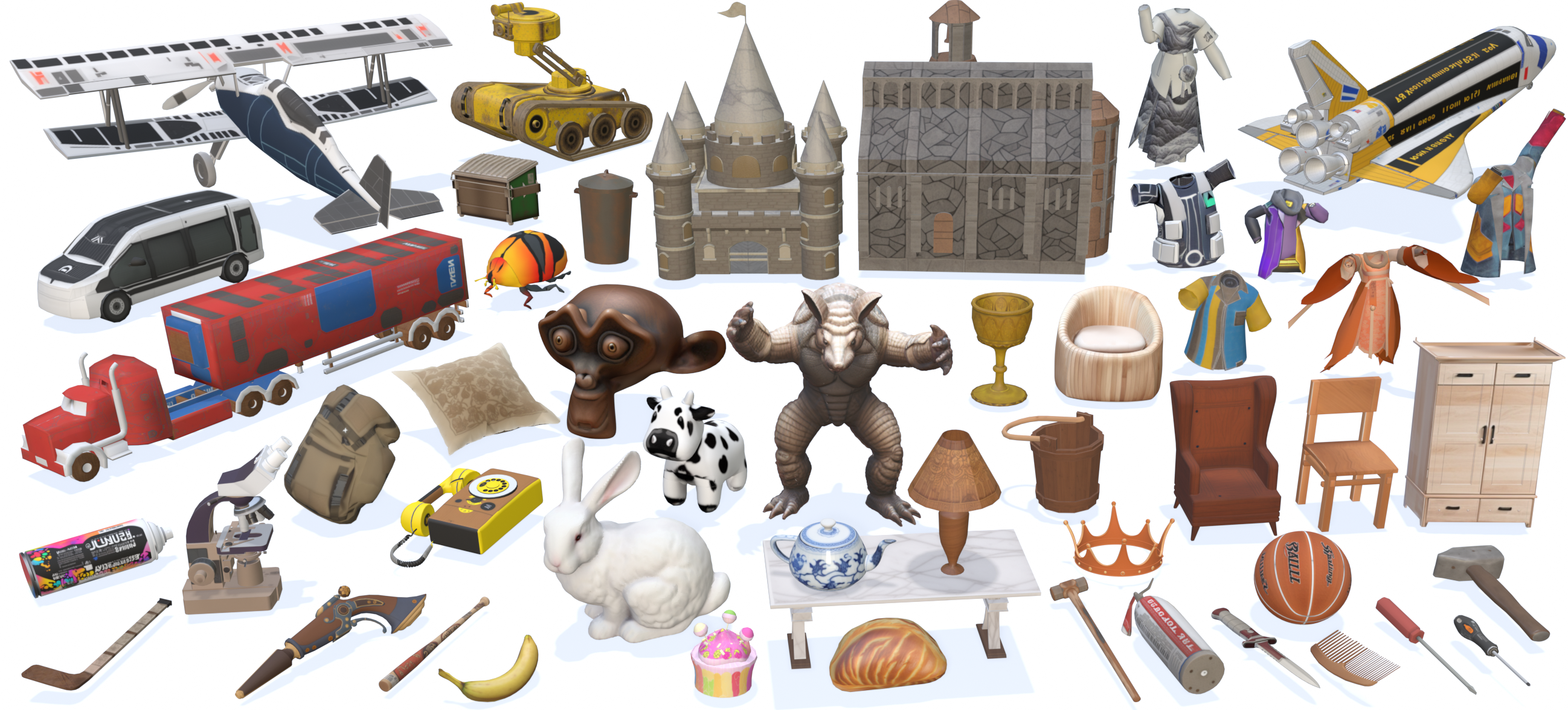}
   \vskip -0.5em
     \captionof{figure}{
     A gallery of generated texture results by Paint3D. 
     Our method is capable of generating lighting-less, high-quality, and high-fidelity textures across diverse objects from numerous categories.
    }
    \label{fig:teaser}
    \bigskip}
\makeatother

\maketitle

\footnotetext[1]{These authors contributed equally to this work.}
\footnotetext[2]{Corresponding author (email: skicyyu@tencent.com).}
%%%%%%%%% ABSTRACT
\begin{abstract}
\vspace{-5pt}
% \vspace{100}
This paper presents Paint3D, a novel coarse-to-fine generative framework that is capable of producing high-resolution, lighting-less, and diverse 2K UV texture maps for untextured 3D meshes conditioned on text or image inputs. The key challenge addressed is generating high-quality textures without embedded illumination information, which allows the textures to be re-lighted or re-edited within modern graphics pipelines. To achieve this, our method first leverages a pre-trained depth-aware 2D diffusion model to generate view-conditional images and perform multi-view texture fusion, producing an initial coarse texture map. However, as 2D models cannot fully represent 3D shapes and disable lighting effects, the coarse texture map exhibits incomplete areas and illumination artifacts. To resolve this, we train separate UV Inpainting and UVHD diffusion models specialized for the shape-aware refinement of incomplete areas and the removal of illumination artifacts. Through this coarse-to-fine process, Paint3D can produce high-quality 2K UV textures that maintain semantic consistency while being lighting-less, significantly advancing the state-of-the-art in texturing 3D objects.
\end{abstract}

%%%%%%%%% BODY TEXT

\section{Introduction}
\label{intro}
The rise of deep generative models has ushered the era of Artificial Intelligence Generated Content, catalyzing advancements in natural language generation~\cite{touvron2023llama,zheng2023judging,2020t5}, image synthesis~\cite{rombach2022high,ramesh2022hierarchical,saharia2022photorealistic,podell2023sdxl}, and 3D generation~\cite{poole2022dreamfusion,wang2023prolificdreamer,liu2023zero}.
These 3D generative technologies have significantly impacted various applications, revolutionizing the landscape of current 3D productions.
However, the generated meshes, characterized by chaotic lighting textures and complex wiring, are often incompatible with traditional rendering pipelines, such as physically based rendering (PBR).
The lighting-less texture diffusion model, capable of generating diverse appearances of 3D assets, should augment these pre-existing 3D productions for the gaming industry, film industry, virtual reality, and so on.

\begin{figure}[t]
    \centering
    \includegraphics[width=\linewidth]{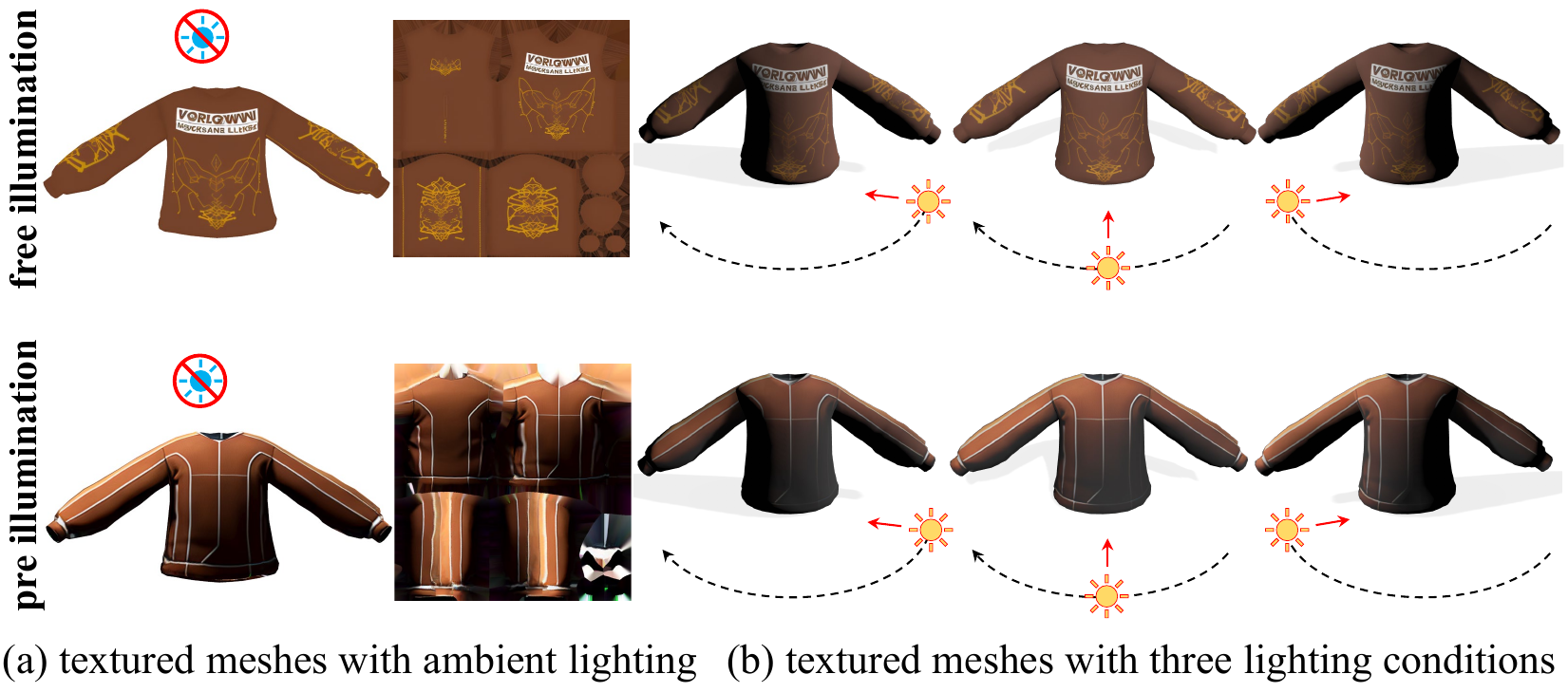}
    \vspace{-17pt}
    \caption{
    Illustration of the pre-illumination problem. 
    The texture map with free illumination is compatible with traditional rendering pipelines,
    while there are inappropriate shadows when relighting is applied on the pre-illumination texture.
    }
    \vspace{-7pt}
    \label{fig:lighiting_changes}
\end{figure}

% (2) Recent methods and their shortages.
Recent advancements in texture synthesis have shown significant progress, particularly in the utilization of 2D diffusion models such as TEXTure~\cite{TEXTure} and Text2tex~\cite{Chen_2023_ICCV}. 
These models effectively employ pre-trained depth-to-image diffusion models to generate high-quality textures through text conditions.
However, these methods have issues with pre-illuminated textures. This can damage the quality of final renderings in 3D environments and cause lighting errors when changing lighting within common graphics workflows, as shown in the bottom of Fig.~\ref{fig:lighiting_changes}.
Conversely, texture generation methods trained from 3D data offer an alternative approach such as PointUV~\cite{yu2023texture} and Mesh2tex~\cite{Bokhovkin_2023_ICCV}, which typically generate textures by comprehending the entire geometries for specific 3D objects.
However, they are often hindered by a lack of generalization, struggling to apply their models to a broad range of 3D objects beyond their training datasets, as well as generate various textures through different textual or visual prompts.
%
% Our model, Paint3D, addresses these limitations by achieving a remarkable level of generalization across diverse 3D models and objects. Unlike previous methods, Paint3D operates effectively without the need for pre-illumination, thereby ensuring the integrity and consistency of textures across various lighting conditions and viewing angles."
%

Two challenges are crucial for texture generation. 
The first is achieving broad generalization across various objects using diverse prompts or image guidance,
and the second is eliminating the coupled illumination on the generated results obtained from pre-training.
Recent advancement of conditioned image synthesis works~\cite{zhang2023adding,rombach2022high} using billion-level images, capable of ``rendering'' diverse image results from 3D views, can help overcome the size limitation of 3D data in texture generation.
However, the pre-illuminated textures can interfere with the final visual outcomes of these textured objects within rendering engines.
Furthermore, since the pre-trained image diffusion models only provide 2D results in the view domain, they struggle to maintain view consistency for 3D objects due to the lack of comprehensive understanding of their shapes.
Therefore, our focus is on developing a two-stage texture diffusion model for 3D objects. This model should be able to generalize to various pre-trained image generative models and learn lighting-less texture generation while preserving view consistency.

% (4)  our method
In this work, we propose a coarse-to-fine texture generation framework, namely Paint3D, that leverages the strong image generation and prompt guidance abilities of pre-trained image generative models for texturing 3D objects.
% overall
To enable the generalization of rich and high-quality texture results from diverse prompts, we first progressively sample multi-view images from a pre-trained view depth-aware 2D image diffusion model and then back-project these images onto the surface of the 3D mesh to generate an initial texture map.
% coarse stage
In the second stage, Paint3D focuses on generating lighting-less textures. To achieve this, we contribute separate UV Inpainting and UVHD diffusion models specialized in the shape-aware refinement of incomplete regions and removal of lighting influences. We train these diffusion models on UV texture space, using feasible 3D objects and their high-quality illumination-free textures as supervision. Through this coarse-to-fine process, Paint3D can generate semantically consistent high-quality 2K textures devoid of intrinsic illumination effects. Extensive experiments demonstrate that Paint3D achieves state-of-the-art performance in texturing 3D objects with different texts or images as conditional inputs and offers compelling advantages for graphics editing and synthesis tasks.

% %
% We train a diffusion model on the texture domain, using feasible 3D objects and their high-quality textures as supervision.
% % fine stage
% By performing this diffusion process in the UV space, Paint3D enhances the completeness and visual aesthetics of the coarse texture maps and further generates lighting-less texture maps.
% % image-to-texture
% Moreover, \cx{Paint3D utilizes xx for image conditions xxx.}
% % summarize
% Extensive experiments demonstrate that Paint3D achieves state-of-the-art performance in texturing 3D objects, while the replaceable image generative module allows Paint3D to adapt to the existing strong text-to-image or image-to-image models, for enriching the capabilities of 3D assets.

% (6)  contributiuons
We summarize our contributions as follows: 
% whole framework
1) We propose a novel coarse-to-fine generative framework that is capable of producing high-resolution, lighting-less, and diverse 2K UV texture maps for untextured 3D meshes;
% refiner
2) We separately design a shape-aware UV Inpainting diffusion model and a shape-aware UVHD diffusion model as the refinement of incomplete regions and removal of lighting influences;
% results
3) Our proposed Paint3D supports both textual and visual prompts as conditional inputs and achieves state-of-the-art performance on texturing 3D objects. The code will be released later.

\begin{figure*}[t]
    \centering
    \includegraphics[width=\linewidth]{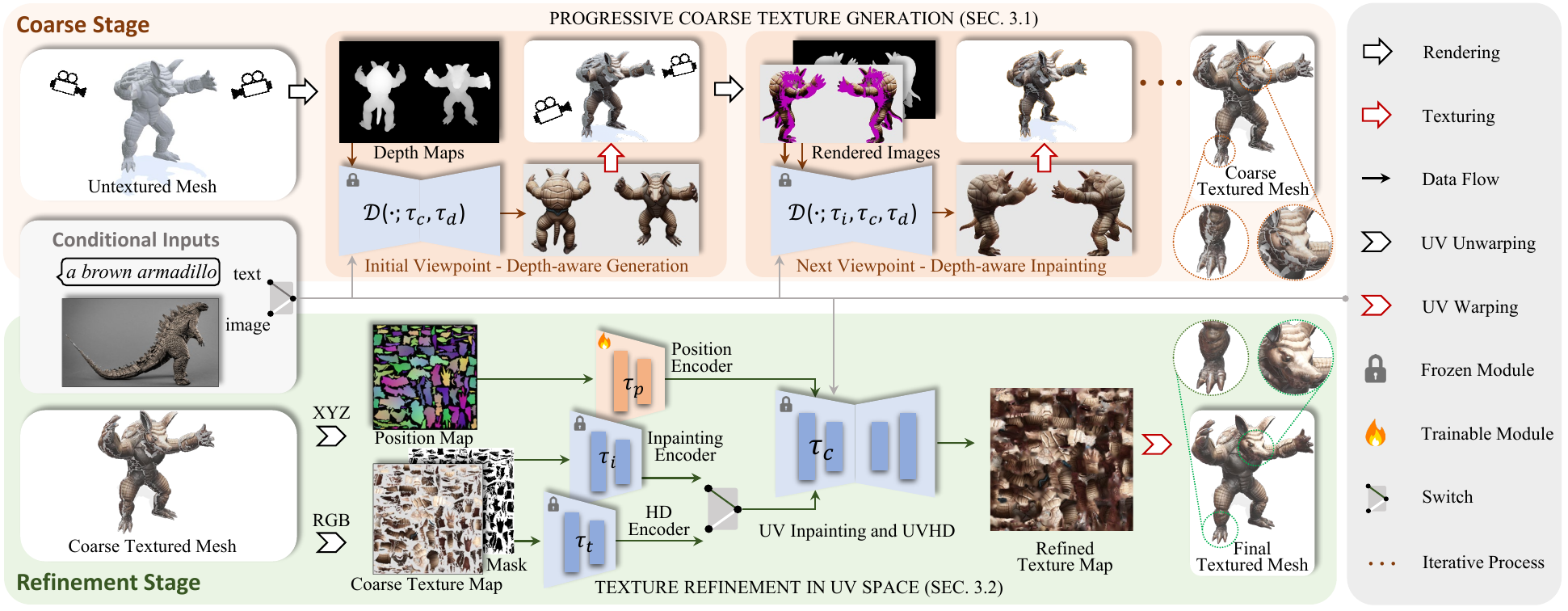}
    \vspace{-17pt}
    \caption{
    The overview of our coarse-to-fine framework. 
    The coarse stage (\cref{sec:method:coarse_stage}) samples multi-view images from the pre-trained 2D image diffusion models, then back-projects these images onto the mesh surface to create initial texture maps. 
    The refinement stage (\cref{sec:method:fine_stage}) generates high-quality textures with a diffusion model in UV space, conditioned on the position map and the coarse texture map.
    }
    \vspace{-7pt}
    \label{fig:pipeline}
\end{figure*}

\section{Related Work}
\label{relatedwork}

Traditional methods~\cite{wei2009state,kopf2007solid,lefebvre2006appearance,turk2001texture,wei2001texture,zhou2014color,huang2020adversarial} of synthesizing texture to 3D assets concentrated on placing simple exemplar patterns on a surface or levering global optimization for painting the 3D shape. However, the recent learning-based approaches~\cite{karnewar2023holofusion,tang2023make,raj2023dreambooth3d,zhuang2023dreameditor,yang2023dreamspace,pan2023enhancing,li2023focaldreamer,chen2023it3d,qian2023magic123} have succeeded in generating plausible textures for more complex 3D shapes. The following discusses the related learning-based methods. 
% categorized into two groups.

% \textbf{Texture Synthesis via 2D Diffusion models.}
\textbf{Iteratively Texturing via 2D Diffusion Models.}
The rapidly expanding large-scale 2D text-to-image (T2I) diffusion models~\cite{rombach2022high,ramesh2022hierarchical,saharia2022photorealistic} have yielded remarkable outcomes, and subsequently, \cite{liu2023zero,tang2023mvdiffusion,shi2023mvdream,long2023wonder3d,li2023sweetdreamer} harness the capabilities of T2I models to facilitate texture synthesis on 3D assets.
% Iteratively Texturing
% Texture, text2tex, texfusion
TEXTure~\cite{TEXTure} devises an iteratively texturing scheme and succeeds in synthesizing high-quality texture. It leverages a pretrained depth-to-image diffusion model and gradually paints the texture map of a 3D model from multiple viewpoints. Although TEXTure~\cite{TEXTure} samples a partial texture map under each viewpoint conditioned on previous results, the generative process still lacks global information modeling, leading to the view-inconsistency results. Later, TexFusion~\cite{cao2023texfusion} proposes to aggregate texture information from different viewpoints during the denoising process and synthesize the entire texture map, which improves the view consistency. Besides, Text2tex~\cite{Chen_2023_ICCV} developed an automatic method to select viewpoints for saving human efforts. These methods improve the global texture modeling but still suffer from the inherited lighting bias from 2D Priors, leading to inconsistent results. 
% In contrast, our model benefits from the 3D data fine-tuning, significantly alleviating the illumination effects.
In contrast, our framework involves a texture refinement model trained with illumination-free texture data, significantly alleviating the illumination artifacts.
% Text2tex\cite{Chen_2023_ICCV} and TexFusion~\cite{cao2023texfusion} 

% Optimization-based Texturing
% Latent Nerf, Compositional 3D, fantasia 3D
\textbf{Optimization-based 3D Generation via 2D diffusion model.} 
Prior to the emergence of large-scale text-to-image models, early 
optimization-based texturing approaches~\cite{michel2022text2mesh,lei2022tango,mohammad2022clip,ma2023x,hong2022avatarclip} endeavored to utilize the large-scale vision-language model, CLIP~\cite{Radford2021LearningTV}, for optimizing texture map of 3D models.
Subsequently, the introduction of Score Distillation Sampling (SDS) in DreamFusion~\cite{poole2022dreamfusion} has paved the way for numerous text-to-3D approaches~\cite{lin2023magic3d,metzer2023latent,chen2023fantasia3d,wang2023prolificdreamer,tsalicoglou2023textmesh,chen2023text,tang2023dreamgaussian,sun2023dreamcraft3d}.
%
% Subsequently, introducing the Score Distillation Sampling (SDS) in the DreamFusion~\cite{poole2022dreamfusion} leads to many text-to-3d approaches ~\cite{lin2023magic3d,metzer2023latent,chen2023fantasia3d,wang2023prolificdreamer,tsalicoglou2023textmesh,chen2023text,tang2023dreamgaussian,sun2023dreamcraft3d}. 
Latent-nerf~\cite{metzer2023latent} and Fantasia3D~\cite{chen2023fantasia3d} extend SDS for optimizing the texture map with texture-less 3D shapes as input. 
Those methods consider inputting an initial shape and simultaneously optimize the texture map and geometry. 
They could produce multi-view consistent texture but cannot guarantee geometry fidelity. Moreover, they struggle with the Janus problem due to the semantically ambiguous.
Different from these methods, our model learns on the whole texture map, preserving the 3D geometry.
% \textcolor{red}{our model learns on the whole texture map, preserving the 3D geometries and synthesizing semantically consistent results.}

\textbf{Generative Texturing from 3D Data.}
 % Texture field, texturify, mesh2tex, TUVF
Various learning-based approaches usually train generative texturing models based on the 3D data~\cite{nichol2022point,luo2023scalable,jun2023shap,li20223d,collins2022abo,deitke2023objaverse} from scratch. 
% According to the texture map representation, we categorize these methods into three types. 
Early methods~\cite{oechsle2019texture,chen2023shaddr,gao2022get3d,gupta20233dgen} learn implicit texture fields to assign a color to each pixel on the surface of the 3D shape. 
However, since the texture on the surface of 3D shapes is continuous, discrete supervision is unlikely to train a model for synthesizing high-quality textures. Texturify~\cite{siddiqui2022texturify} defines texture maps on the surface of polygon meshes and devises a convolution operator for mesh structures by incorporating SytleGAN~\cite{karras2019style,karras2020analyzing,karras2021alias} architecture for predicting texture on each face. 
Such methods are limited by the mesh resolution and the lack of global information modeling, although the recent Mesh2tex~\cite{Bokhovkin_2023_ICCV} further integrates an implicit texture field branch for improvements. Moreover, some methods (AUV-net~\cite{chen2022auv}, LTG~\cite{yu2021learning}, TUVF~\cite{cheng2023tuvf}, PointUV~\cite{yu2023texture}) learn to synthesize UV-Maps for 3D shapes, avoiding the abovementioned limitations. 
Unfortunately, these methods usually struggle when handling more general objects due to the variations between 3D objects in different categories.

% \newpage

\section{Method}
\label{method}
\iffalse
% Preliminary - denoise diffusion models
paragraph 1: overview
paragraph 2: notation
Coarse stage
  - init sample, multi view sample and inverse render
  - depth-aware inpaint
Fine stage
  - UV challage
  - UV pos
  - UVinpaint + UVtile
\fi

% === objective and overview
% In the coarse stage, we sample multi-view images from the pre-trained 2D image diffusion models, and then back-project these images to the mesh surface to get a coarse texture map.
%
To synthesize high-quality and diverse texture maps for 3D models based on desired conditional inputs like prompts or images, we propose a coarse-to-fine framework, Paint3D, which progressively generates and refines texture maps, as shown in~\cref{fig:pipeline}.
In \textbf{the coarse stage} (see~\cref{sec:method:coarse_stage}), we sample multi-view images from the pre-trained 2D image diffusion models, then back-project these images onto the mesh surface to create initial texture maps.
In \textbf{the refinement stage} (see~\cref{sec:method:fine_stage}), we enhance coarse texture maps by performing a diffusion process in the UV space, 
achieving lighting-less, inpainting, High Definition (HD) functions to ensure the final texture's completeness and visual appeal.

% === notation
% condition inputs
% $c_t$
% $c_i$
Given an uncolored 3D model $M$ and an appearance condition $c$, such as text prompts~\cite{TEXTure,Chen_2023_ICCV} or an appearance reference image~\cite{Bokhovkin_2023_ICCV},
% even a empty condition $c = \varnothing$~\cite{siddiqui2022texturify}, 
our Paint3D aims to generate the texture map $T$ for the 3D model.
Here, we represent the 3D model's geometry using a surfaced mesh, denoted as $M = (V,F)$, with vertices $V = \{v_i\}, v_i \in \mathbb{R}^3$ and triangular faces $F = \{f_i\}$, where each $f_i$ is a triplet of vertices.
The texture map is represented by a multi-channel image in UV space, denoted as $T \in \mathbb{R}^{H\times W\times C}$. 
The proposed Paint3D framework $\mathcal{P}$  consists of two stages: the coarse texture generation stage $\mathcal{C}: (M, c) \mapsto \hat{T}$ and the texture refinement stage $\mathcal{F}: \hat{T} \mapsto T$, that is $ T = \mathcal{P}(M, c) = \mathcal{F}(\mathcal{C}(M, c))$.
Furthermore, we define a conditional diffusion model as $\mathcal{D}(\cdot ;\tau_\theta)$, where $\tau_\theta$ is a domain-specific encoder and can be substituted for varying conditions.
% we introduce a domain specific encoder τθ that projects y

\subsection{Progressive Coarse Texture Generation}
\label{sec:method:coarse_stage}
\iffalse
    - render, initial sample, inverse render
    - depth-aware inpaint for non-initial sample
    - the shortage of coarse texture map
\fi

% Overview the course stage
In this state, we generate a coarse UV texture map for untextured 3D meshes based on a pre-trained view depth-aware 2D diffusion model. Specifically, we first render the depth map from different camera views, then sample images from the image diffusion model with depth conditions, and finally back-project these images onto the mesh surface.
% Given a 3D mesh and a conditional diffusion model in the image domain, we can paint the surface appearance of the 3D mesh by rendering the depth map from different camera views, sampling images from the image diffusion model with depth condition, and then back-projecting these images onto the mesh surface.
%
To improve the consistency of textured meshes in each view, we alternately perform the three processes of rendering, sampling, and back-projection, progressively generating the entire texture map~\cite{TEXTure,Chen_2023_ICCV}.

% render, initial sample, inverse render
% and then sample texture image $I_1$ from an image diffusion model according to depth condition and appearance-related condition, denoted as $\mathcal{D}_{I}:(z, d_1, c) \mapsto I_1$.
\textbf{Initial Viewpoint.} 
With the set of camera views $\{p_i\}_{i=1}^n$ focusing on the 3D mesh, we start to generate the texture of the visible region.
We first render the 3D mesh to a depth map $d_1$ from the first view $p_1$, where this rendering process is denoted as $\mathcal{R}:(M, p_1) \mapsto d_1$. 
% from a geometry-aware image diffusion model,
We then sample a texture image $I_1$ given an appearance condition $c$ and a depth condition $d_1$,  denoted as
\begin{equation}
\label[equation]{eq:depth_generation}
I_1 = \mathcal{D}(z, c, d_1;\tau_c, \tau_d),
\end{equation}
where $z \in \mathbb{R}^{h\times w \times e}$ is a random initialized latent, $\tau_c$ is appearance encoder, and $\tau_d$ is depth encoder.
% Given the appearance-related condition $c$ and the depth condition $d_1$, 
% we then sample an appearance image $I_1$ from an image diffusion model with a random latent $z \in \mathbb{R}^{h\times w \times c}$, 
% denoted as $\mathcal{D}:(c, d_1, z ) \mapsto I_1$.
%
Subsequently, we back-project this image onto the 3D mesh from the first view, generating the initial texture map $\hat{T}_1$, 
where this back-projecting process is denoted as $\mathcal{R}^{-1}:(M, I_1, p_1) \mapsto \hat{T}_1$.

% render, inpaint sample, inverse render
% The combination of xxx and xxx, paint mesh from the first view.
\textbf{Next Non-initial Viewpoint.} For these viewpoints $p_k$, we execute a similar process as mentioned above but the texture sampling process is performed in an image inpainting manner.
% render
Specifically, taking into account the textured region from all previous viewpoints $\hat{T}_{\{1,k-1\}}$, the rendering process outputs not only a depth image $d_k$ but also a partially colored RGB image $\hat{I}_k$ and an uncolored area mask $m_k$ in the current view, 
denoted as $\mathcal{R}:(M, p_k, \hat{T}_{\{1,k-1\}} ) \mapsto (d_k, \hat{I}_k, m_k)$.
We use a depth-aware image inpainting model, with a new inpainting encoder $\tau_i$, to fill the uncolored area within the rendered RGB image, denoted as 
\begin{equation}
\label[equation]{eq:depth_inpainting}
I_k = \mathcal{D}(\hat{I}_k, m_k, c, d_k;\tau_i, \tau_c, \tau_d).
\end{equation}
% $\mathcal{D}:(c, d_k, \hat{I}_k, m_k) \mapsto {I}_k $.
%
The inpainted image is back-projected onto the 3D mesh under the current view, generating the current texture map $\hat{T}_k$ from the view $p_k$,
denoted as $\mathcal{R}^{-1}:(M, I_k, p_k) \mapsto \hat{T}_k$.
The textured region from previous viewpoints $\hat{T}_{\{1,k-1\}}$ is kept and the uncolored area is updated by the current texture map $\hat{T}_k$, formatted as
\begin{equation}
\hat{T}_{\{1,k\}} = m_{k-1}^{UV} \odot \hat{T}_{\{1,k-1\}} + (1- m_{k-1}^{UV}) \odot \hat{T}_k,
\end{equation}
where $m_{k-1}^{UV}$ is the colored area mask in the UV plane and can be calculated from the texture map $\hat{T}_{\{1,k-1\}}$.
Therefore, the texture map is progressively generated view-by-view and arrives at the entire coarse texture map $\hat{T} = \hat{T}_{\{1,n\}}$.
%
% % 
% We provide more details about the architecture
% and the training in the supplementary
% %
% We also explore the effectiveness
% of the latent’s dimensions on motion sequences representation in Tab. 4. Hence, our VAE models present a stronger
% motion reconstruction ability and richer diversity (cf. Tab. 5
% and Tab. 6). We provide more details about the architecture
% and the training in the supplementary

\textbf{Multi-view Texture Sampling.}
We extend the texture sampling process mentioned above (\cref{eq:depth_generation} and \cref{eq:depth_inpainting}) to the multi-view scene.
Specifically, in the initial texture sampling, we utilize a pair of cameras to capture two depth maps $\{d_1, d_2\}$ from symmetric viewpoints.
We then concatenate those two depth maps horizontally (in width) and compose a depth grid with a size of $1\times 2$, denoted as $\mathbf{d_1}$.
To perform multi-view depth-aware texture sampling, we replace the single depth image $d_1$ with the depth grid $\mathbf{d_1}$ in~\cref{eq:depth_generation}.
Similarly, in the non-initial texturing, we horizontally concatenate renders, composing depth grid $\mathbf{d_k}$, RGB image grid $\mathbf{\hat{I}_k}$, and mask grid $\mathbf{m_k}$.
To perform multi-view depth-aware texture inpainting, we replace the inputs in~\cref{eq:depth_inpainting} with those grids.
As evaluated in \cref{sec:exp:ablation}, we also explore the effectiveness of the number of viewpoints.

\subsection{Texture Refinement in UV Space}
\label{sec:method:fine_stage}
\iffalse
  - coarse map remains an xxx problem: lighting, hole and enhance visual
  - challange in UV space
  - diffusion in UV space + light-less
  - UVinpaint and UVtile
\fi

% === coarse map remains xxx problem: lighting, hole and enhance visual
Although the appearance of the coarse texture map is coherent, it still has some issues like lighting shadows involved by the 2D image diffusion model, or the texture holes caused by self-occlusion during the rendering process.
We propose to perform a diffusion process in the UV space based on the coarse texture map, aiming to mitigate these issues and further enhance the visual aesthetics of the texture map during texture refinement.
% To mitigate these issues, we propose conducting a diffusion process in the UV space based on the coarse texture map. 
% Moreover, refining the textures allows for a more realistic and intricate appearance, improving the overall visual quality and perception.
%
% === Challenge in UV space, inconsistency
% However, performing the diffusion process in the UV space with mainstream 2D image diffusion models~\cite{} presents the challenge of texture discontinuity~\cite{Yu_2023_ICCV}.
However, refining texture maps in the UV space with mainstream image diffusion models~\cite{rombach2022high,Zhang_2023_ICCV} presents the challenge of texture discontinuity~\cite{yu2023texture}.
% why inconsistency
% Indeed,
The texture map is derived through UV mapping of the 3D surface texture, which cuts the continuous texture on the 3D mesh into a series of individual texture fragments in the UV plane.
This fragmentation complicates the learning of the 3D adjacency relationships among the fragments in the UV plane, 
leading to texture discontinuity issues.
% This fragmentation introduces a significant challenge for the diffusion model, as it complicates the learning of the 3D adjacency relationships among the fragments in the UV plane.
% leading to discontinuity issues when the generated texture map is applied to the 3D mesh surface.
% Consequently, this can lead to discontinuity issues when the generated texture map is applied to the 3D mesh surface.

% === position map
\textbf{Position Encoder.} To refine the texture map in UV space, we perform the diffusion process guided by adjacency information of texture fragments.
Here, the 3D adjacency information of texture fragments is represented as the position map in UV space $O \in \mathbb{R}^{H\times W\times 3}$, where each non-background element 
% $O_{u,v}=(x,y,z)$ 
is a 3D point coordinate.
Similar to the texture map, the position map can be obtained through UV mapping of the 3D point coordinates.
To fuse the 3D adjacency information during the diffusion process, we add an individual position map encoder $\tau_p$ to the pretrained image diffusion model. 
Following the design principle of ControlNet~\cite{Zhang_2023_ICCV}, the new encoder has the same architecture as the encoder in the image diffusion model and is connected to it through zero-convolution layer.
% $O_{u,v}=(x,y,z), O_{u,v} \notin \varnothing $ 
% The parameters of image diffusion models are fixed and the 3D adjacency information is fused via 'zero convolutions', aiming at reusing the powerful generation ability of image diffusion models

% === training
Our texture diffusion model is trained using a dataset consisting of paired position maps and texture maps $\{O_i, T_i\}_{i=1}^n$.
Given a set of conditions including time step $t$, appearance condition $c$, as well as a position map $O$
, our texture diffusion model learns to predict the noise added to the noisy latent $z_t$ with
\begin{equation}
    \label[equation]{eq:loss:con}
    \mathcal{L} =\mathbb{E}_{z_0, t, c, O ,\epsilon \sim \mathcal{N}(0,1)}\left[\left\|\epsilon-\epsilon_\theta\left(z_t, t, c,  \tau_p(O)\right)\right\|_2^2\right].
\end{equation}
% given
For an image diffusion model with a trained denoiser $\epsilon_\theta$, we freeze $\epsilon_\theta$ as suggested by~\cite{Zhang_2023_ICCV} and only optimize the position encoder $\tau_p$ with~\cref{eq:loss:con}.
Since texture maps in UV space are lighting-less, our model can learn this prior from data distribution, generating lighting-less texture.

% UVinpaint and UVtile
\textbf{UV Inpainting.} We can simultaneously use the position encoder and other conditional encoders to perform various refinement tasks in UV space.
% Here we introduce two specific refinement capabilities, texture inpainting in UV space (UV inpainting) and texture tiling in UV space (UV-tiling).
Here we introduce two specific refinement capabilities, namely UV inpainting and UV High Definition (UVHD).
% Suppose $\mathcal{D}(\cdot ; \Theta)$ is a trained image diffusion model, with parameters $\Theta$, 
% that transforms an input condition $c$ into a target image output $I$, as
% $I = \mathcal{D}(c, z; \Theta)$.
%
The UV inpainting is used to fill texture holes within the UV plane, which can avoid self-occlusion problems during rendering.
% To achieve UV inpainting, we concurrently utilize the positional map encoder $\tau_p$  and an image inpainting encoder $\tau_i$ with a trained diffusion model $\mathcal{D}$, denoted as
To achieve UV inpainting, we add the position map encoder $\tau_p$ on an image inpainting diffusion model as
% \tilde{T}
\begin{equation}
T_{inpainting} = \mathcal{D}(\hat{T}, m^{UV}, c, O;\tau_i, \tau_c, \tau_p),
\end{equation}
which takes as input a coarse texture map $\hat{T}$, texture map mask $m^{UV}$, appearance condition $c$, and position map $O$, and produces as output an inpainted texture map $T_{inpaint}$.
 
\textbf{UV High Definition (UVHD)} is designed to enhance the visual aesthetics of the texture map.
We use the position encoder $\tau_p$  and an image enhance encoder $\tau_t$ with a diffusion model $\mathcal{D}(\cdot; \tau_c)$ to achieve UVHD, denoted as
\begin{equation}
T_{tiling} = \mathcal{D}(\hat{T}, c, O ;\tau_t, \tau_c, \tau_p).
\end{equation}
% The UV-tiling takes as input an appearance condition $c$, position map $O$, coarse texture map $\hat{T}$, and produces as output a tiled texture map $T_{tiling}$.
In our refinement stage, we perform UV inpainting followed by UVHD to get the final refined texture map $T$.
% \begin{equation}
% T = \mathcal{D}(c, O, \hat{T}, m^{UV} ;\tau_p, \tau_i),
% \end{equation}
% 
% We provide more details about the architecture in the supplementary.
%
By integrating the UV inpainting and UVHD, Paint3D is capable of producing lighting-less (\cref{fig:ablation:fine}), complete (\cref{fig:ablation:UVinpaint}), high-resolution, and diverse UV texture maps (\cref{fig:ablation:UVHD}). 
% We also explore the effectiveness of UV inpainting and UVHD during the texture refining process (details in \cref{sec:exp:ablation}).

% \newpage
\section{Experiments}
\label{experiments}

We provide extensive comparisons to evaluate our models on both quality and diversity in the following. 
Firstly, we introduce the datasets settings, evaluation metrics and
implementation details~\cref{sec:exp:details}. 
Importantly, we show the comparisons on two texture generation tasks, including text-to-texture
(\cref{sec:exp:cmp_text}), image-to-texture (\cref{sec:exp:cmp_img}).
Lastly, we conduct ablation studies to demonstrate the effectiveness of each module in our Paint3D (\cref{sec:exp:ablation}).
More qualitative results, comparisons, and details are provided in supplements.

% details + data + baseline + metrics
\subsection{Implementation Details}
\label{sec:exp:details}

% text , image, ControlNet
We apply the text2image model from Stable Diffusion v1.5~\cite{rombach2022high} as our texture generation backbone.
To handle the image condition, we employ the image encoder introduced in IP-Adapter~\cite{ye2023ip-adapter}.
For additional conditional controls such as depth, image inpainting, and image high definition, we utilize the domain encoders provided in ControlNet~\cite{Zhang_2023_ICCV}.
In the coarse texture generation, we define six axis-aligned principal viewpoints, and sample two texture images from a pair of symmetric viewpoints during a single diffusion progress.
The denoising strengths are set as 1 and 0.75 for the coarse and refinement stages, respectively. 
Our implementation uses the PyTorch~\cite{paszke2017automatic} framework, with Kaolin~\cite{KaolinLibrary} used for rendering and texture projection.
For the UV unwarping process, we utilize the original UV map if the mesh contains texture coordinates, or we use an open-source UV-Atlas tool~\cite{JonathanYoung18} to perform UV unwarping.

\textbf{Datasets.}
% dataset and number
We conduct experiments on a subset of textured meshes from the Objaverse~\cite{deitke2023objaverse} dataset.
% To ensure the quality of the texture map, 
We exclude meshes devoid of textures, those with monochromatic texture, and 3D scene objects composed of multiple meshes.
The filtered subset contains 105,301 texture meshes, with 105,000 meshes utilized for training the position encoder and 301 meshes employed for evaluating our model.
Additionally, we gather 30 meshes in the wild to assess our model. This brings the total to 331 high-quality textured meshes for evaluation.

\textbf{Evaluation metrics.}
We access the generated textures with commonly used metrics for image quality and diversity. 
Specifically, we report the Frechet Inception Distance (FID)~\cite{heusel2017gans} and Kernel Inception Distance (KID $\times 10^{-3}$)~\cite{BinkowskiSAG18}. 
To calculate the generated image distribution, we render 512 $\times$ 512 images of each mesh with the synthesized textures, captured from 20 fixed viewpoints. 
The real distribution is made up of renders of the meshes under identical settings, but using their original textures.

\begin{figure*}[t]
    \centering
    \includegraphics[width=\linewidth]{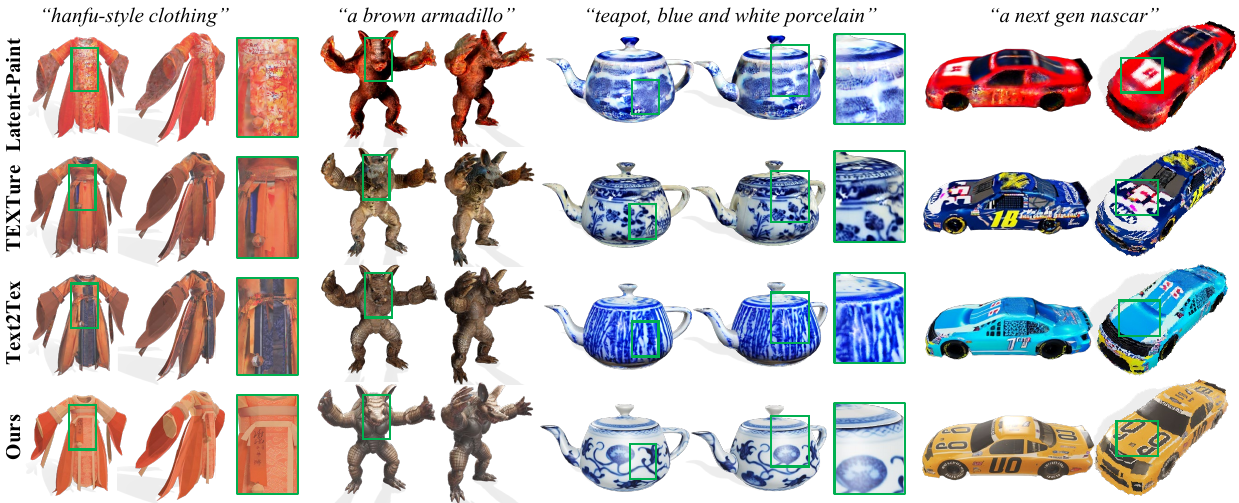}
    \vspace{-17pt}
    \caption{
    Qualitative comparisons on texture generation conditioned on text prompt. 
    We compare our textured mesh against Latent-Paint~\cite{metzer2023latent}, TEXTure~\cite{TEXTure}, and Text2Tex~\cite{Chen_2023_ICCV}.
    Compared to the baselines, our method generates an illumination-free texture map, as well as more exquisite texture details ($cf.$ supplements for more our results). 
    % Please zoom in for details 
    }
    \vspace{-5pt}
    \label{fig:compare}
\end{figure*}

% what - why (optional) - baseline - details -results
\subsection{Comparisons on Text-to-Texture}
\label{sec:exp:cmp_text}
We first evaluate the texture generation effect of Paint3D conditioned on the text prompt.
We compare our method with state-of-the-art approaches, including Latent-Paint~\cite{metzer2023latent}, TEXTure~\cite{TEXTure}, and Text2Tex~\cite{Chen_2023_ICCV}.
%
% Latent-Paint explicitly manipulates the texture map via the txt2image model from Stable Diffusion, 
% It has the capability to generate textures on any 3D mesh.
Latent-Paint is a texture generation variant of the NeRF-based 3D object generation framework, 
% specifically designed for texture generation. 
explicitly manipulating the texture map via the text2image model from Stable Diffusion. 
TEXTure devises an iterative texture generation scheme to manipulate the texture map, and successfully synthesizes high-quality textures.
Following a similar principle, Text2Tex develops an automatic viewpoint selection strategy in the iterative process, representing the current state-of-the-art in the field of text-conditioned texture generation.
For the category-specific texture generation approaches~\cite{yu2023texture,siddiqui2022texturify,Bokhovkin_2023_ICCV}, we provide more comparisons in the supplements.

\begin{table}[t]
\centering
% \small
% \tabcolsep=2\tabcolsep
\resizebox{\columnwidth}{!}{
\begin{tabular}{@{}lccccc@{}}
\toprule
\multirow{2}{*}{Methods} & \multicolumn{1}{c}{\multirow{2}{*}{{FID$\downarrow$}}} & \multicolumn{1}{c}{\multirow{2}{*}{KID  $\downarrow$}} & \multicolumn{2}{c}{User Study} \\
 & \multicolumn{1}{c}{} & \multicolumn{1}{c}{} & \multicolumn{1}{c}{Overall Quality$\uparrow$} & \multicolumn{1}{c}{Text Fidelity$\uparrow$} \\
% Methods              & {FID$\downarrow$}  & {KID  $\downarrow$} & {Overall Quality$\uparrow$} & {Text Fidelity$\uparrow$} \\ 
\hline
% \toprule
Latent-Paint~\cite{metzer2023latent} & 
62.22 &  15.81   & 2.83   & 3.29  \\ 

TEXTure~\cite{TEXTure} & 
43.13 & 11.13  & 3.36   & 4.12  \\ 

Text2Tex~\cite{Chen_2023_ICCV} & 
38.93 & 7.94  & 3.57   & 4.27  \\ 

Ours  & $\boldsymbol{27.28}$ & $\boldsymbol{4.81}$  & $\boldsymbol{4.45}$   & $\boldsymbol{4.74}$  \\ 
\bottomrule
\end{tabular}%
}
\vspace{-8pt}
\caption{Quantitative comparisons on text-to-texture task. Ours outperforms other approaches on both FID and KID ($\times 10^{-3}$).}
\label{tab:compare:text}
\vspace{-10pt}
\end{table}

\textbf{Qualitative comparisons}.
% Each row is the result of one method and the textured mesh is displayed from two perspectives for each text prompt.
As shown in~\cref{fig:compare}, our approach is able to generate an illumination-free texture map while excelling at synthesizing high-quality texture details.
Firstly, Latent-Paint~\cite{metzer2023latent} tends to generate blurry textures, which can lead to suboptimal visual effects.
Additionally, while TEXTure~\cite{TEXTure} is capable of generating clear textures, the generated textures may lack smoothness and exhibit noticeable seams or splicing(e.g., the teapot in~\cref{fig:compare}).
Lastly, even though Text2Tex~\cite{Chen_2023_ICCV} demonstrates the ability to generate smoother textures, it may compromise in generating fine textures with intricate details.
Notably, all baselines generate pre-illumination texture maps that led to inappropriate shadows when relighting was applied.

\textbf{Quantitative comparisons}.
In~\cref{tab:compare:text}, we present the quantitative comparisons with the previous SOTA methods in text-driven texture synthesis.
% demonstrating that our method surpasses existing works.
%
Following~\cite{Chen_2023_ICCV,yu2023texture}, we report the FID~\cite{heusel2017gans} and KID~\cite{BinkowskiSAG18} to access the quality and diversity of the generated texture maps. 
Our method outperforms all baselines by a significant margin (29.93\% improvement in FID and 39.42\% improvement in KID). 
These improvements demonstrate the superior capability of our method in generating high-quality textures across diverse objects from numerous categories.

\textbf{User study}.
We further conduct a user study to analyze the overall quality of the generated textures and their fidelity to the input text prompts.
We randomly select 60 meshes and corresponding text prompts to perform the user study. 
Those meshes are textured by both Paint3d and baseline models, and displayed to users in random sequence.
Each object displays full-view texture details in the form of 360-degree rotation. 
Each respondent is asked to evaluate the results based on two aspects: (1) overall quality and (2) fidelity to the text prompt, using a scale of 1 to 5.
We collected the evaluation results of 30 users, as presented in~\cref{tab:compare:text}, where we show the average results across all prompts for each method. 
As can be seen, our approach outperforms all baselines in terms of both overall quality and text fidelity by a significant margin.

\begin{figure}[t]
    \centering
    \includegraphics[width=\linewidth]{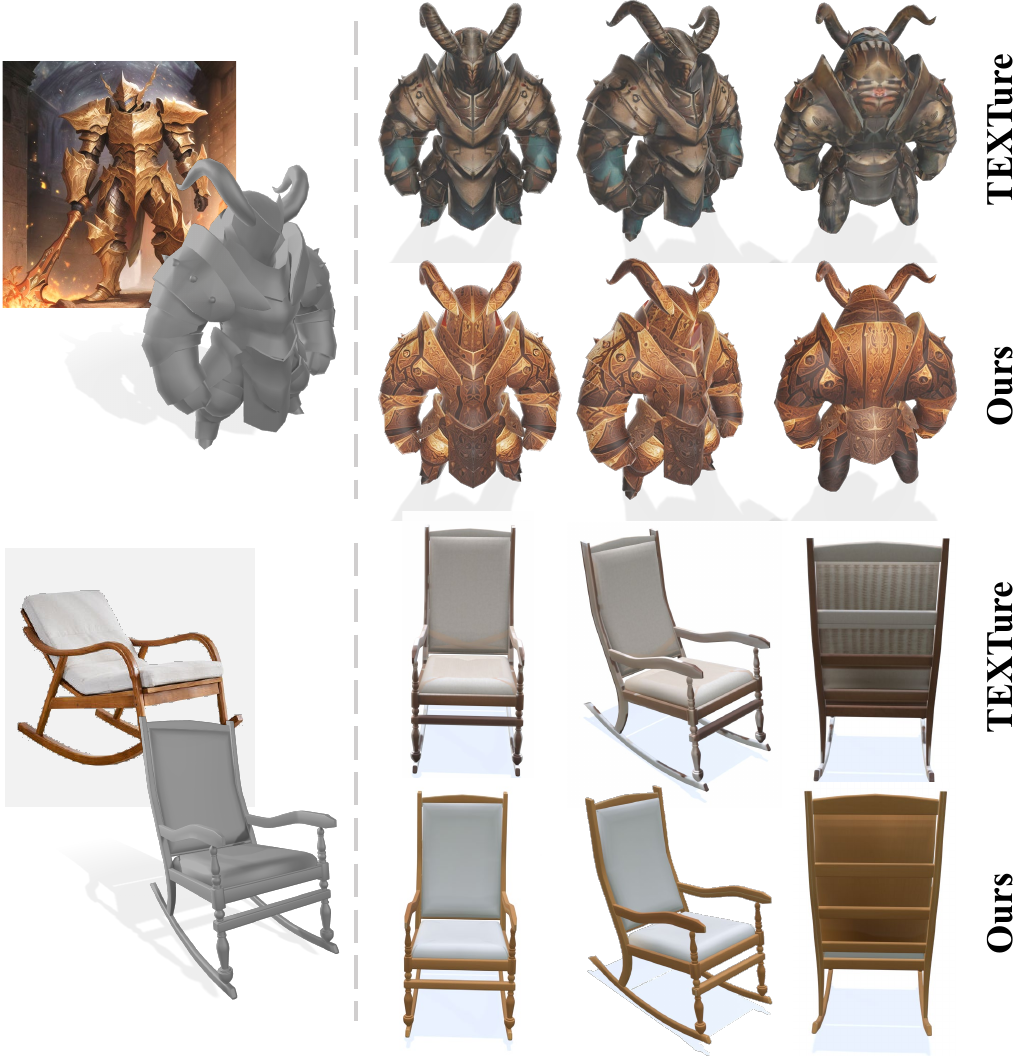}
    \vspace{-17pt}
    \caption{
    Qualitative comparisons on texture generation conditioned on image prompt.
    Compared to TEXTure, our method can better represent texture details contained in the image condition. 
    }
    \vspace{-7pt}
    \label{fig:compare:img}
\end{figure}

\begin{table}[t]
\centering
% \small
% \tabcolsep=2\tabcolsep
\resizebox{\columnwidth}{!}{
\begin{tabular}{@{}lccccc@{}}
\toprule
\multirow{2}{*}{Methods} & \multicolumn{1}{c}{\multirow{2}{*}{{FID$\downarrow$}}} & \multicolumn{1}{c}{\multirow{2}{*}{KID  $\downarrow$}} & \multicolumn{2}{c}{User Study} \\
 & \multicolumn{1}{c}{} & \multicolumn{1}{c}{} & \multicolumn{1}{c}{Overall Quality$\uparrow$} & \multicolumn{1}{c}{Image Fidelity$\uparrow$} \\
\hline
% \toprule

TEXTure~\cite{TEXTure} & 
40.83 & 9.76  & 3.56   & 3.73  \\  

Ours  & $\boldsymbol{26.86}$ & $\boldsymbol{4.94}$  & $\boldsymbol{4.71}$   & $\boldsymbol{4.89}$  \\ 
\bottomrule
\end{tabular}%
}
\vspace{-8pt}
\caption{
Quantitative comparisons on image-to-texture task. 
Our method achieves a significant improvement over the baseline.
% Ours outperforms the baseline on FID and KID ($\times 10^{-3}$). 
}
\label{tab:compare:img}
% \vspace{-10pt}
\end{table}

\subsection{Comparisons on Image-to-Texture}
\label{sec:exp:cmp_img}
We then evaluate the texture generation capability of Paint3D conditioned on the image prompt.
Here, we provide TEXTure~\cite{TEXTure} as our comparison baseline.
% Texture details
We use the texture transfer capability of TEXTure to generate its image-to-texture results.
% Specifically, we provide the input image condition as a 
% Our details
% To generate textures conditioned on the image input, 
To handle the image condition,
our Paint3D employs the image encoder introduced in~\cite{ye2023ip-adapter} based on the txt2image model from Stable Diffusion v1.5~\cite{rombach2022high}.
As depicted in~\cref{fig:compare:img}, our approach excels in synthesizing exquisite texture while maintaining high fidelity with respect to the image condition.
TEXTure~\cite{TEXTure} is capable of generating a similar texture as the input image, but it struggles to accurately represent texture details in the image condition.
For instance, in the samurai case, TEXTure generates a golden armor texture but fails to synthesize high-frequency line details present on the armor.

As shown in~\cref{tab:compare:img}, we also report the FID~\cite{heusel2017gans} and KID~\cite{BinkowskiSAG18} scores under the image condition.
Our method demonstrates a significant improvement over the baseline, as evidenced by the FID score decreasing from 40.83 to 26.86 and the KID score decreasing from 9.76 to 4.94.
% with the FID score improving from 40.83 to 26.86 and the KID score improving from 9.76 to 4.94.
For the user study, we follow a similar evaluation setting as described in ~\cref{sec:exp:cmp_text}, but replace the text prompt with the image prompt.
Each participant needs to assess the generated texture based on its overall quality and fidelity to the image prompt, using a rating scale ranging from 1 to 5.
The average scores of all users are reported in~\cref{tab:compare:img}.
Notably, Paint3D gets a 4.89 average score on image fidelity, indicating our method is able to accurately represent texture details contained in the image condition.

\begin{table}[t]
\centering
% \small
% \tabcolsep=2\tabcolsep
\resizebox{\columnwidth}{!}{
\begin{tabular}{@{}cccccc@{}}
\toprule
\multirow{2}{*}{Coarse Stage} & \multicolumn{2}{c}{Refinement Stage} & \multirow{2}{*}{FID$\downarrow$} & \multirow{2}{*}{KID  $\downarrow$} \\ 
 & \multicolumn{1}{c}{UV inpainting} & UVHD &  &  \\
% Coarse Stage              & UV inpainting  & UVHD & {FID$\downarrow$}  & {KID  $\downarrow$}  \\ 
\hline
% \toprule
\checkmark & \crossmark &  \crossmark   & 41.84   & 10.91  \\ 
\crossmark & \checkmark &  \checkmark   & 48.81   & 11.98  \\ 
\checkmark & \checkmark &  \crossmark   & 37.84   & 7.13  \\ 
\checkmark & \crossmark &  \checkmark   & 33.42   & 6.19  \\ 
\checkmark & \checkmark &  \checkmark   & $\boldsymbol{27.28}$   & $\boldsymbol{4.81}$  \\ 

\bottomrule
\end{tabular}%
}
\vspace{-8pt}
\caption{
Evaluation of modules in the Paint3D framework.
This demonstrates the effectiveness of each component, including the coarse stage, UV inpainting, and UVHD.
By integrating the generation prior in the coarse stage and the illumination-free prior in the refinement stage, our full model achieves the optimal result.
}
\label{tab:ablation:framework}
% \vspace{-7pt}
\end{table}

\begin{figure}[t]
    \centering
    \includegraphics[width=\linewidth]{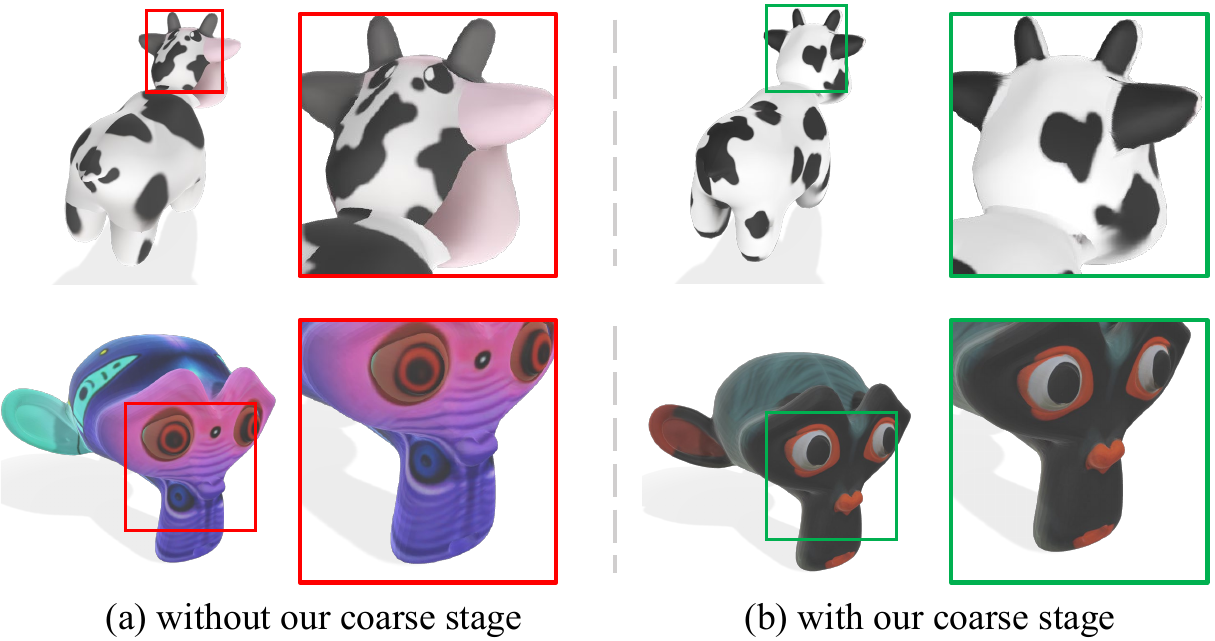}
    \vspace{-17pt}
    \caption{
    Illustration of the effect of the coarse stage.
    % Without our coarse stage, the generated textures have noticeable semantic confusion.
    The absence of our coarse stage may result in semantic confusion in the texture.
    }
    \vspace{-7pt}
    \label{fig:ablation:coarse}
\end{figure}

\begin{figure}[t]
    \centering
    \includegraphics[width=\linewidth]{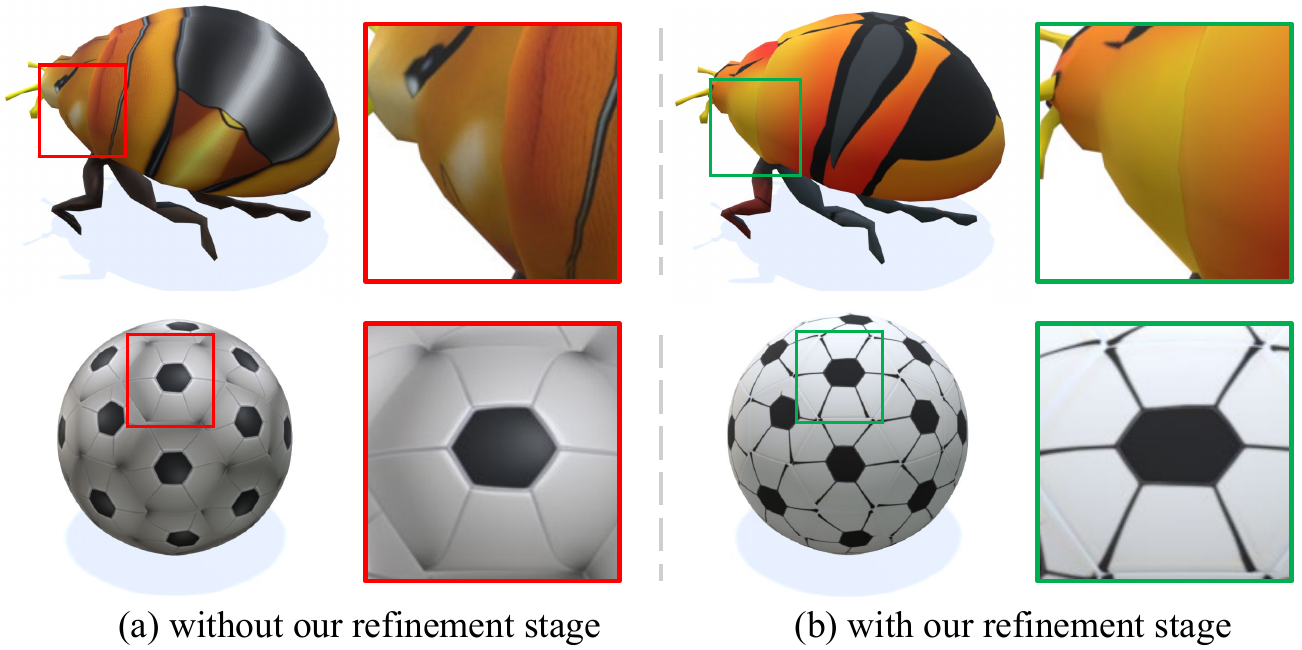}
    \vspace{-17pt}
    \caption{
    Visualization of the effect of the refinement stage.
    With our refinement stage, the generated textures are illumination-free.
    }
    \vspace{-7pt}
    \label{fig:ablation:fine}
\end{figure}

% \subsection{Evaluation on Refinement Stage}
\subsection{Ablation Studies}
\label{sec:exp:ablation}
\textbf{Evaluation of Coarse-to-fine Framework.}
To demonstrate the effectiveness of our coarse-to-fine texture generation framework, we conduct experiments on two baselines ``w/o coarse stage'' and ``w/o refinement stage''. 
The ``w/o coarse stage'' configuration refers to directly generating the texture map using the texture refinement modules in UV space, performing UV inpainting followed by UVHD without initialization from the coarse stage.
The ``w/o refinement stage'' configuration represents the outcome of the coarse stage, where the uncolored area is assigned a color using bilinear interpolation.
In both scenarios, the model produces inferior results compared to our full model, as
reported in~\cref{tab:ablation:framework}. 
We visualize the results of ``w/o coarse stage'' in~\cref{fig:ablation:coarse}.
Absent the coarse stage, the generated textures may display noticeable semantic problems, as the texture map in UV space consists of separate texture fragments. 
% This fragmentation poses a challenge for achieving semantic coherence and understanding.
% 
As shown in in~\cref{fig:ablation:fine}, without the refinement stage, the generated textures are pre-illuminated.

\begin{figure}[t]
    \centering
    \includegraphics[width=\linewidth]{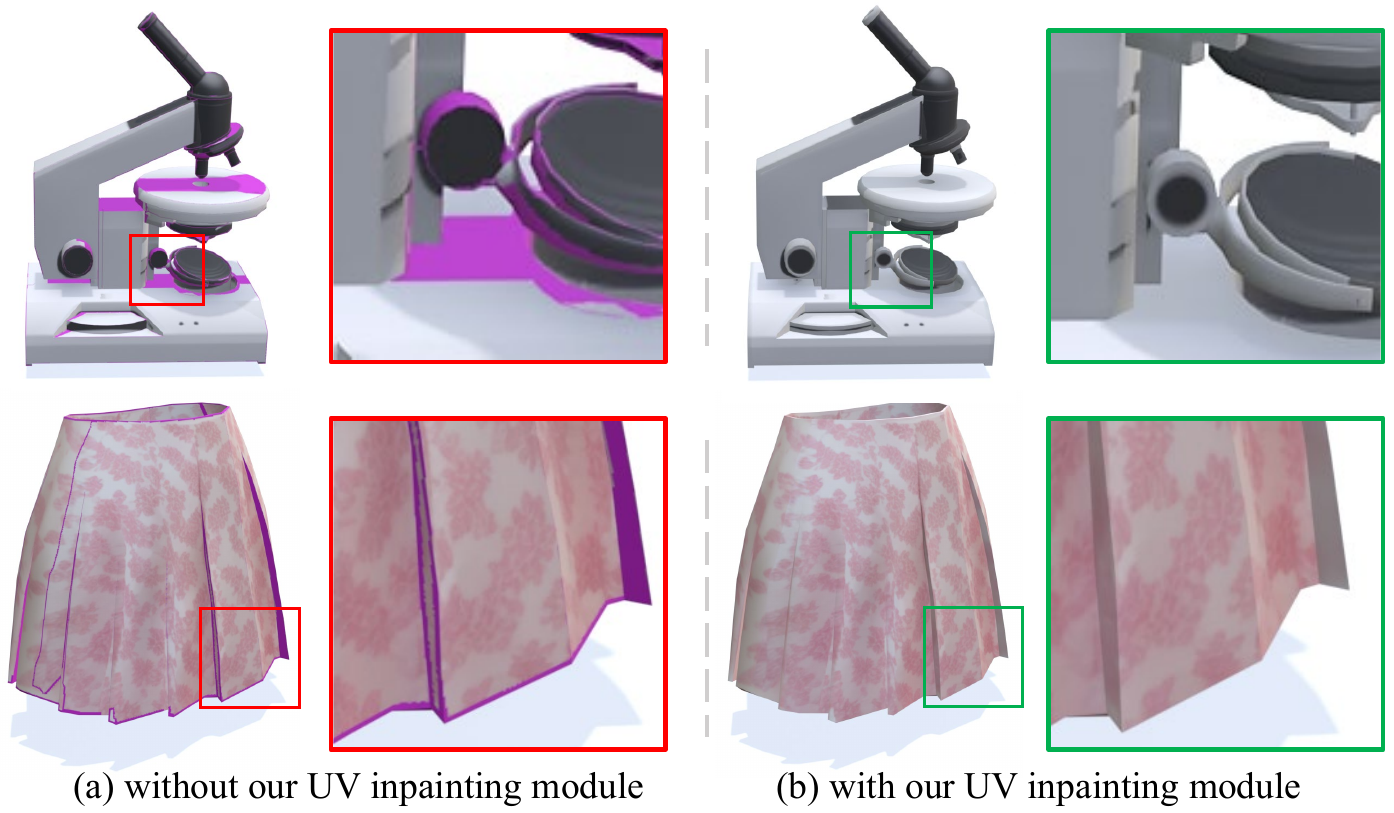}
    \vspace{-17pt}
    \caption{
    Illustration of the effect of UV inpainting.
    UV inpainting can effectively fill texture holes that are located in projecting blind spots (\eg the inner side of a pleated skirt).
    }
    \vspace{-5pt}
    \label{fig:ablation:UVinpaint}
\end{figure}

\begin{figure}[t]
    \centering
    \includegraphics[width=\linewidth]{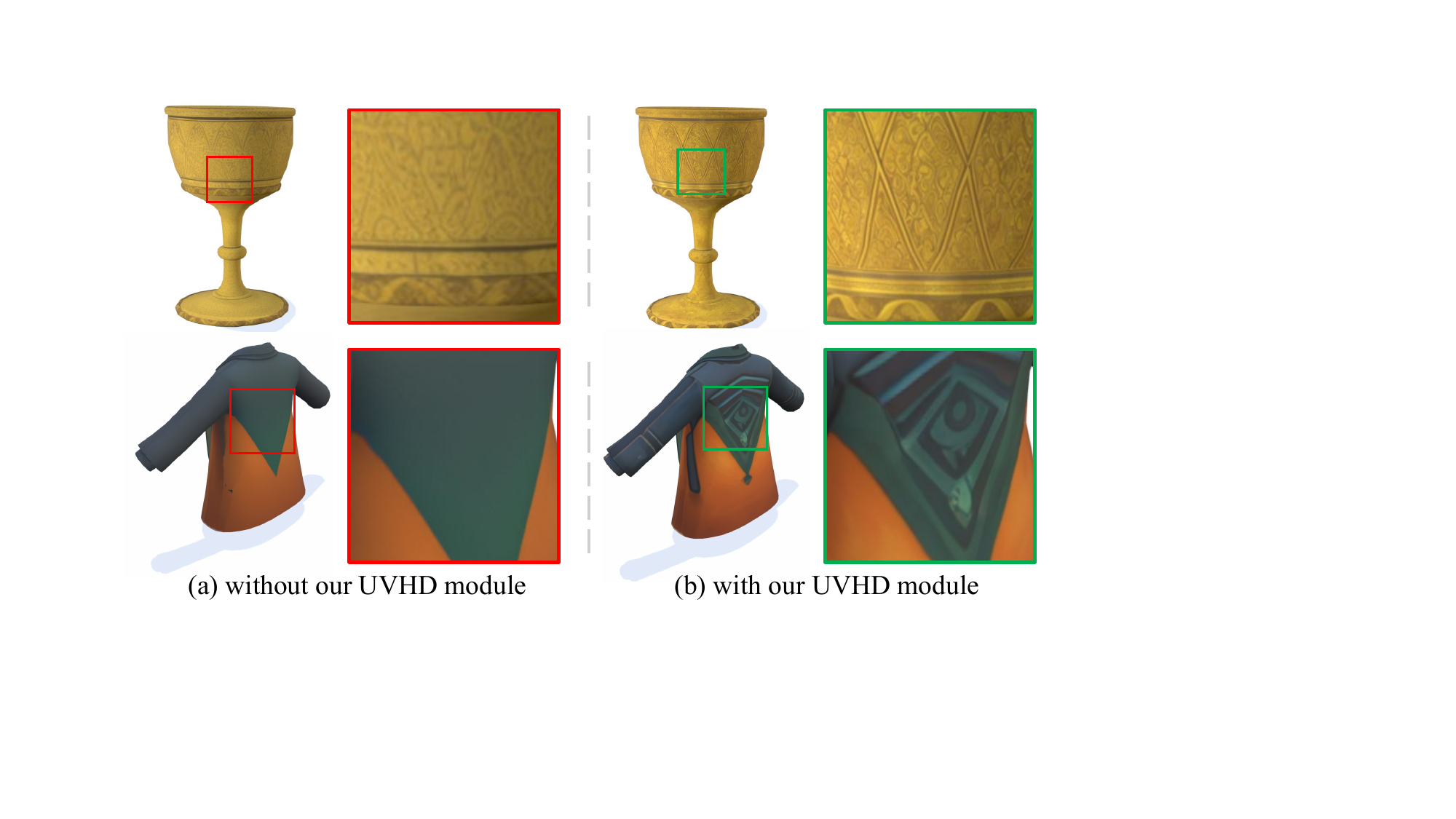}
    \vspace{-17pt}
    \caption{
    Illustration of the effect of UVHD module.
    This displays the capability of UVHD to enhance existing texture details and can even generate new textures in monochromatic areas.
    }
    \vspace{-5pt}
    \label{fig:ablation:UVHD}
\end{figure}

\textbf{Evaluation of UV inpainting and UVHD.}
To demonstrate the effectiveness of two texture refinement modules, UV inpainting and UVHD,
we further conduct experiments on two baselines ``w/o UV inpainting'' and ``w/o UVHD''. 
The ``w/o UV inpainting'' configuration refers to filling the uncolored area with the bilinear interpolation instead of UV inpainting, followed by the UVHD module.
The ``w/o UVHD" configuration represents the inpainted result of the coarse stage with the UV inpainting module.
%
% In both scenarios, the model produces inferior results compared to our full model. 
As indicated in \cref{tab:ablation:framework}, the performance shows a significant decrease when UV inpainting or UVHD is not utilized, indicating their irreplaceable function during texture refinement processing.
We visualize the results of ``w/o UV inpainting'' in~\cref{fig:ablation:UVinpaint}.
UV inpainting can effectively fill texture holes that are located in blind spots, as this inpainting processing is performed within the UV plane, without occlusion problems.
As depicted in Figure \ref{fig:ablation:UVHD}, UVHD demonstrates its capability to enhance exsiting texture details and even generate new textures on monochromatic areas. 
% (as observed in the cloth case).

\textbf{Evaluation of the Number of Viewpoints.}
The selection of viewpoints has shown a significant influence on the texture generation result in the coarse stage~\cite{Chen_2023_ICCV}.
We conduct ablation studies to analyze the impact of the number of viewpoints on both overall coarse texture generation and single diffusion process.
%
% As illustrated in Table \ref{tab:ablation:views}, we observe that increasing the number of viewpoints can enhance the quality of generated textures. However, it is important to note that the relationship between the number of viewpoints and the results is not a simple "more is better" scenario.
As shown in~\cref{tab:ablation:views}, we can see that increasing the number of viewpoints can improve the quality of generated textures, but it is not that the more the viewpoints the better the results.
We achieve the best result when the viewpoint is set to 6.
The result is further improved when we sample two texture images from a pair of symmetric viewpoints during a single diffusion progress.
% We may
% 

\begin{table}[t]
\centering
% \small
% \tabcolsep=2\tabcolsep
\resizebox{\columnwidth}{!}{
\begin{tabular}{cccccccc}
\toprule
\multicolumn{2}{c}{\#Viewpoint} & \multicolumn{1}{c}{\multirow{2}{*}{FID$\downarrow$}} & \multicolumn{1}{c}{\multirow{2}{*}{KID  $\downarrow$}} & \multicolumn{2}{c}{\#Viewpoint} & \multicolumn{1}{c}{\multirow{2}{*}{FID$\downarrow$}} & \multicolumn{1}{c}{\multirow{2}{*}{KID  $\downarrow$}} \\
\multicolumn{1}{c}{Total} & \multicolumn{1}{c}{One Iter} & \multicolumn{1}{c}{} & \multicolumn{1}{c}{} & \multicolumn{1}{c}{Total} & \multicolumn{1}{c}{One Iter} & \multicolumn{1}{c}{} & \multicolumn{1}{c}{} \\
\hline
2 & 1 & 42.31 & 11.67 & 2 & 2 & 41.74 & 10.19 \\
4 & 1 & 36.07 & 7.85 & 4 & 2 & 32.60 & 6.37 \\
6 & 1 & 29.02 & 5.10 & 6 & 2 & $\boldsymbol{27.28}$ & $\boldsymbol{4.81}$ \\
8 & 1 & 30.15 & 5.65 & 8 & 2 & 27.71 & 4.93 \\

\bottomrule
\end{tabular}
}
\vspace{-8pt}
\caption{
Evaluation of the number of viewpoints in the coarse stage. 
The viewpoints are not the more the better, as the pretrained 2D image diffusion model may involve illumination artifacts.
}
\label{tab:ablation:views}
\end{table}

\vspace{5pt}
\section{Disscusion}
This paper presents Paint3D, a novel coarse-to-fine generative framework that is capable of generating high-quality 2K UV textures that maintain semantic consistency while being lighting-less, conditioned on text or image inputs.
To achieve this, our method first leverages a pre-trained depth-aware 2D diffusion model to generate view-conditional images and perform multi-view texture fusion, producing an initial coarse texture map.
Subsequently, we train distinct UV Inpainting and UVHD diffusion models, specifically designed for shape-aware refinement of incomplete areas and the removal of illumination artifacts.
Through this coarse-to-fine process, Paint3D can produce high-quality, lighting-less, and diverse texture maps, significantly advancing the state-of-the-art in texturing 3D objects.

% limitations
Our method has inherent limitations as follows.
% Moreover, our method’s efficacy relies on the quality of UV mapping. 
% the UV mapping produces excessive fragmented cuts, resulting in fragmented artifacts. 
Our approach still suffers from the multi-faces problem in the coarse stage which will result in a failure case. This issue primarily arises from the inconsistency of multi-view texture images sampled by the pre-trained 2D diffusion model, as it is not explicitly trained on multi-view datasets.
% We believe that fine-tuning or retraining 2D diffusion models on large-scale multi-view datasets will improve the multi-view consistency of textures.
%
% Moreover, 
It remains a challenge for Paint3D to generate material maps, which are commonly used in modern physically based rendering pipelines.
Furthermore, unlike optimization-based 3D generation methods~\cite{lin2023magic3d,metzer2023latent,chen2023fantasia3d,wang2023prolificdreamer},
Paint3D is not capable to generate or edit the geometry of 3D assets.

\newpage
{\small
\bibliographystyle{ieee_fullname}
\bibliography{egbib}
}

\onecolumn
\section*{\hfil {\LARGE Appendix}\hfil}
% \vspace{50pt}
% \section*{\hfil {\LARGE Paper ID 150}\hfil}
\renewcommand\thesection{\Alph{section}}
\renewcommand*{\theHsection}{appedix.\thesection}
\setcounter{section}{0}
\setcounter{figure}{9}
\setcounter{table}{4}
\setcounter{equation}{6}

This appendix provides more qualitative results (\cref{sec:appendix:qualitative}), 
several additional experiments (\cref{sec:appendix:exps}), 
and discussion on the failure cases of our proposed texture generation approach (\cref{sec:appendix:failure}).
% evaluations of inference time (\cref{sec:appendix:inferencetime}).
% user study (\cref{sec:appendix:userstudy}).

% \myparagraph{Video.} 
% \textbf{Video.}
% We have provided supplemental videos on the project page ($\cf$ the html file).
% We have provided supplemental videos on the project page.
% In these supplemental videos, we show 1) comparisons of text-based texture generation, 2) comparisons of image-conditional texture generation, and 3) more samples of our texture generation results.
% We suggest the reader watch this video for dynamic results.\\

% \myparagraph{Code} will be released later.
% is available on \href{https://github.com/chenfengye/motion-latent-diffusion}{GitHub Page}. 
% We provide the process of the training and evaluation of MLD models, the pre-trained model files, the demo script, and example results.

\section{Qualitative Results}
\label{sec:appendix:qualitative}

\begin{figure}[h]
    \centering
    \includegraphics[width=\linewidth]{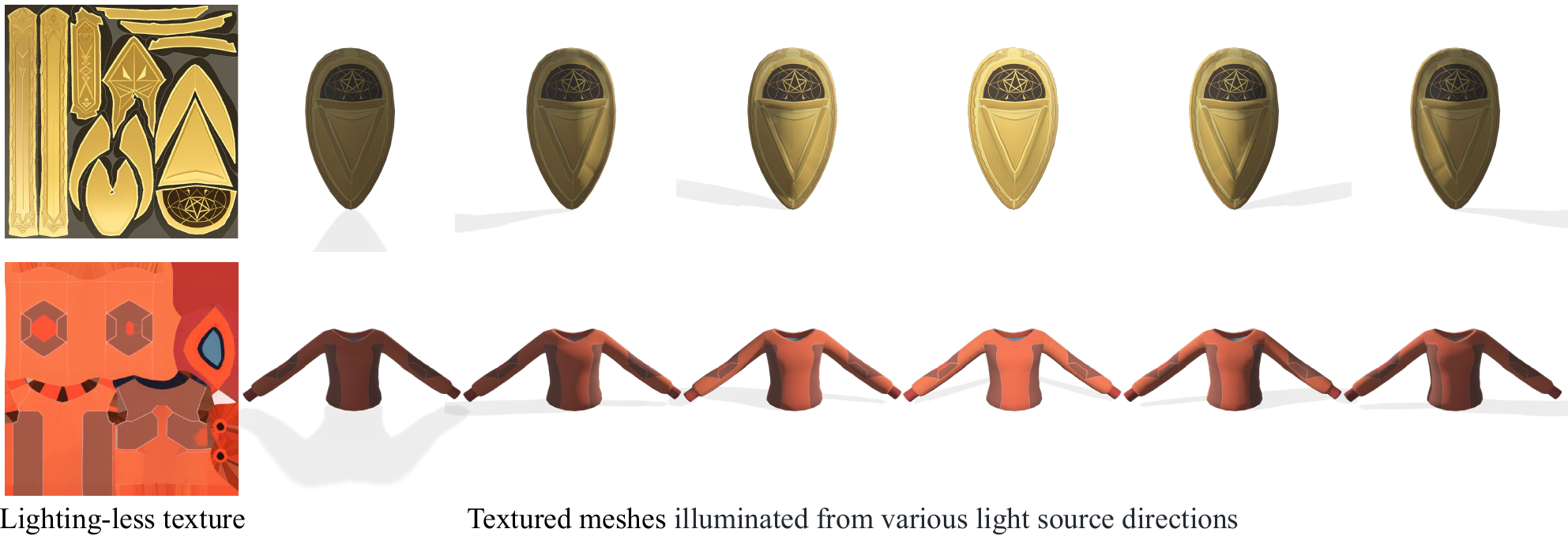}
    \caption{
    Lighting-less texture maps generated by Paint3D.
    These lighting-less textures produce appropriate shadows when the textured meshes are illuminated from different directions of light sources.
    % The lighting-less texture is compatible with traditional rendering pipelines.
    % There are appropriate shadows when textured meshes are illuminated from various light source directions.
    }
    \label{fig:appendix:lighting}
\end{figure}

\begin{figure}[h]
    \centering
    \includegraphics[width=\linewidth]{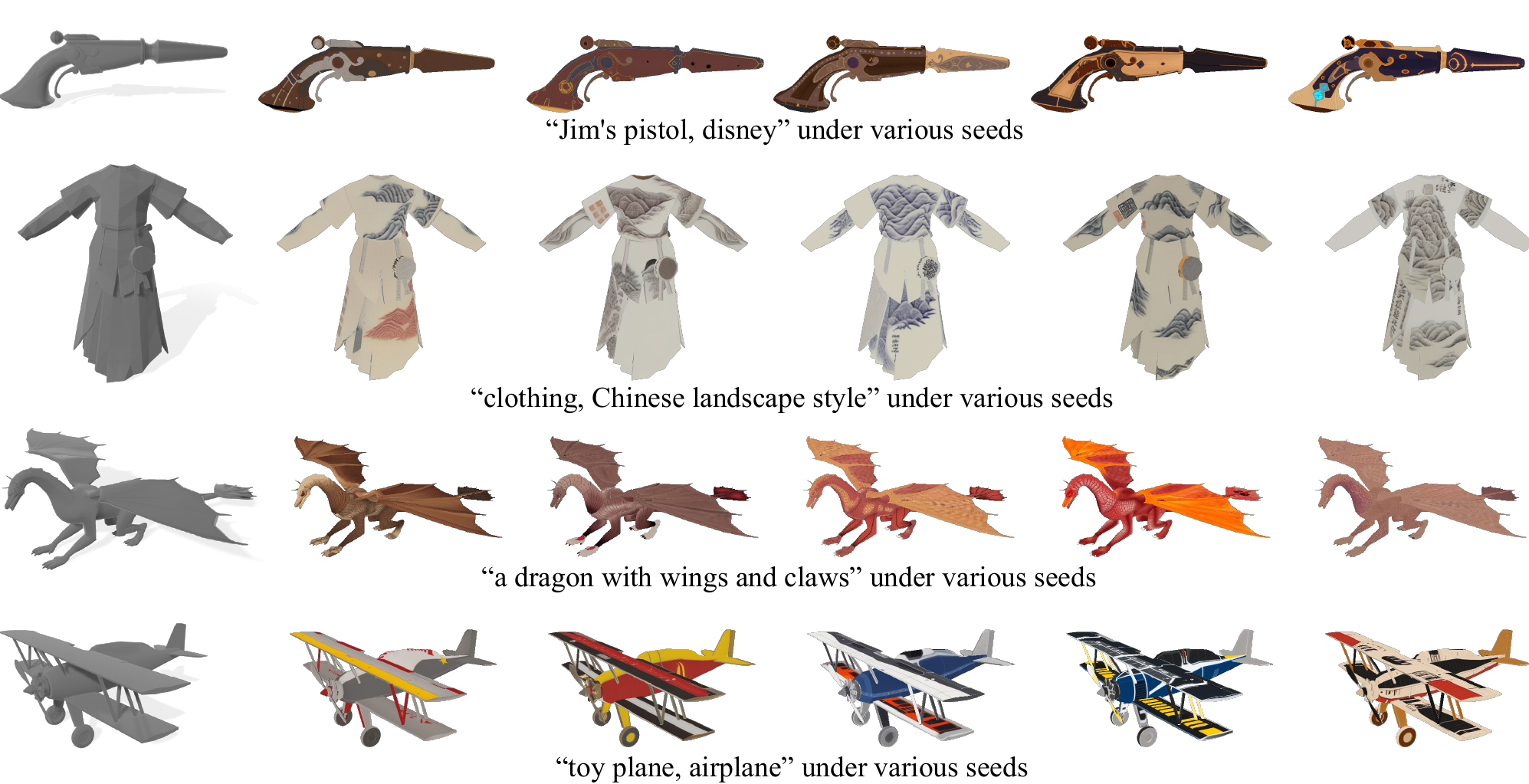}
    \caption{
    More samples from our best model for text-to-texture generation. 
    Samples are generated with text prompts of the test set under various seeds. 
    We recommend the supplemental video to see more results. 
    }
    \label{fig:appendix:text_style}
\end{figure}

\begin{figure}[h]
    \centering
    \includegraphics[width=\linewidth]{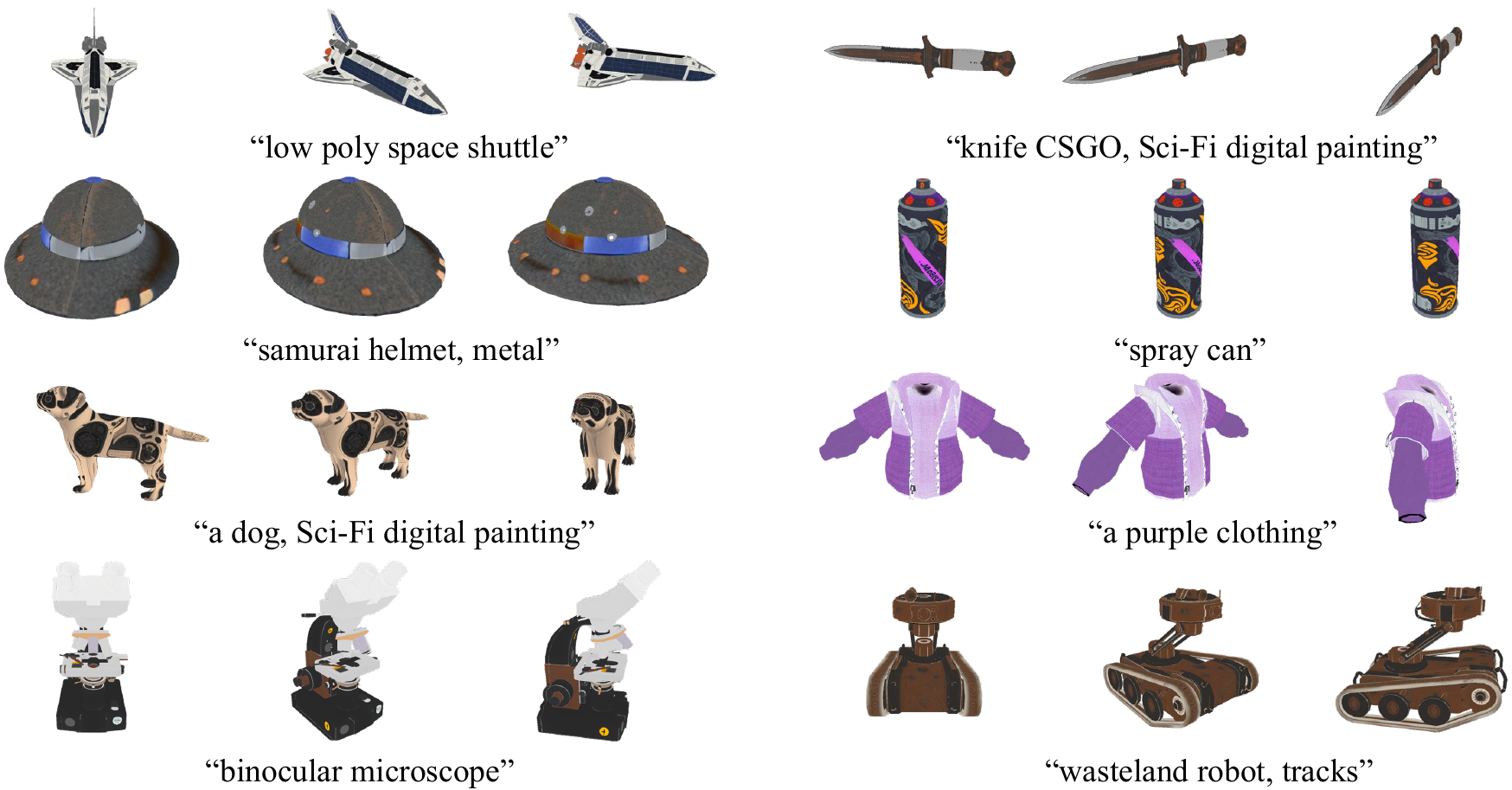}
    \caption{
    Additional texturing results generated by Paint3D on text-to-texture task.
    Each textured mesh is shown from three viewpoints.
    }
    \label{fig:appendix:text}
\end{figure}

\begin{figure}[h]
    \centering
    \includegraphics[width=\linewidth]{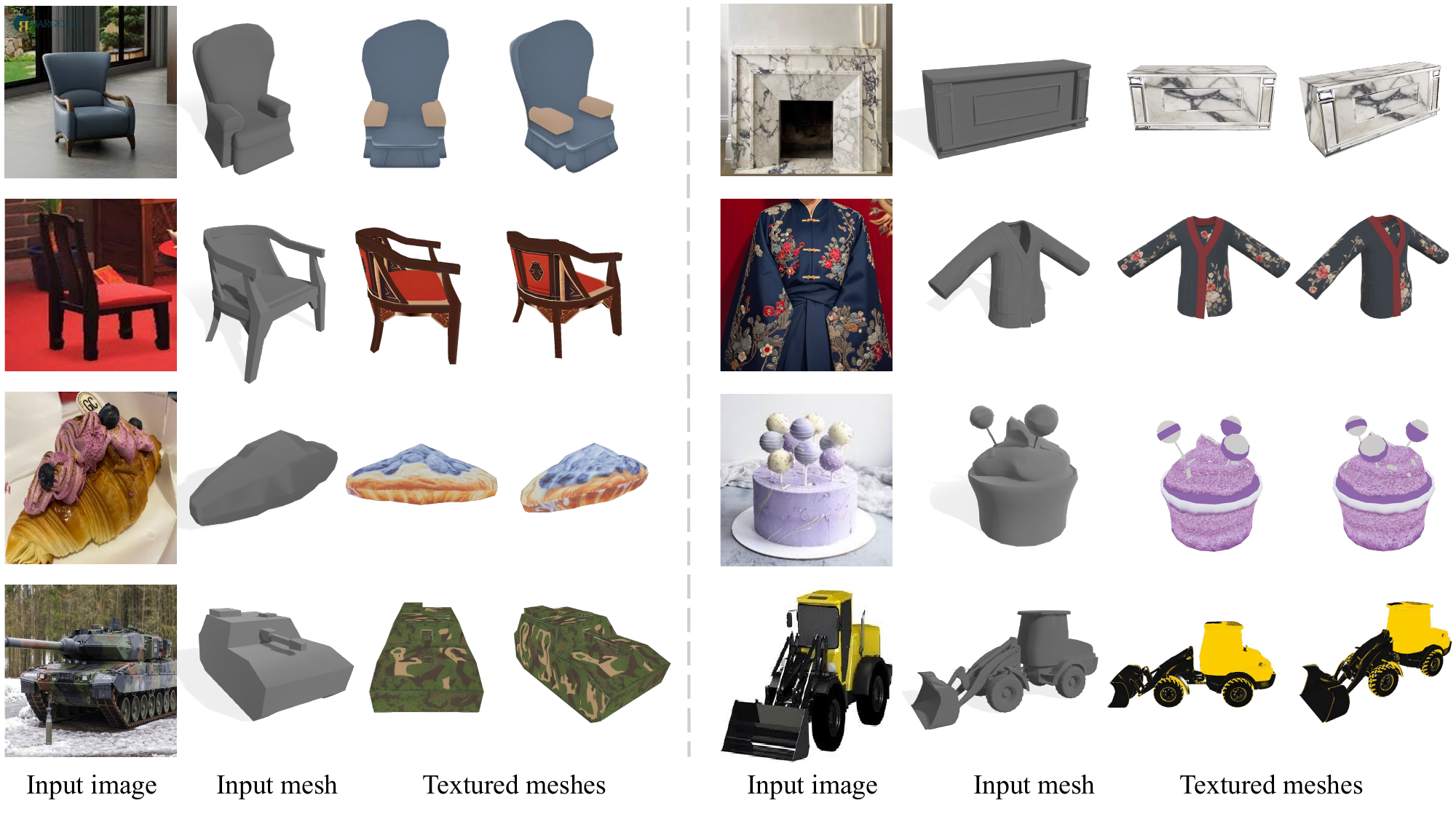}
    \caption{
    Additional samples from Paint3D for image-to-texture generation and each textured mesh is shown from two viewpoints.
    The input image conditions are collected in the wild.
    We recommend the supplemental video to see more results. 
    }
    \label{fig:appendix:image}
\end{figure}

\newpage
\section{Additional Experiments}
\label{sec:appendix:exps}
We first study the effectiveness of the position map in the UV Inpaint and UVHD modules.
Then, we provide more comparisons with category-specific texture generation approaches~\cite{yu2023texture}.

\subsection{Evaluation of Position Map}
\begin{figure}[h]
    \centering
    \includegraphics[width=0.95\linewidth]{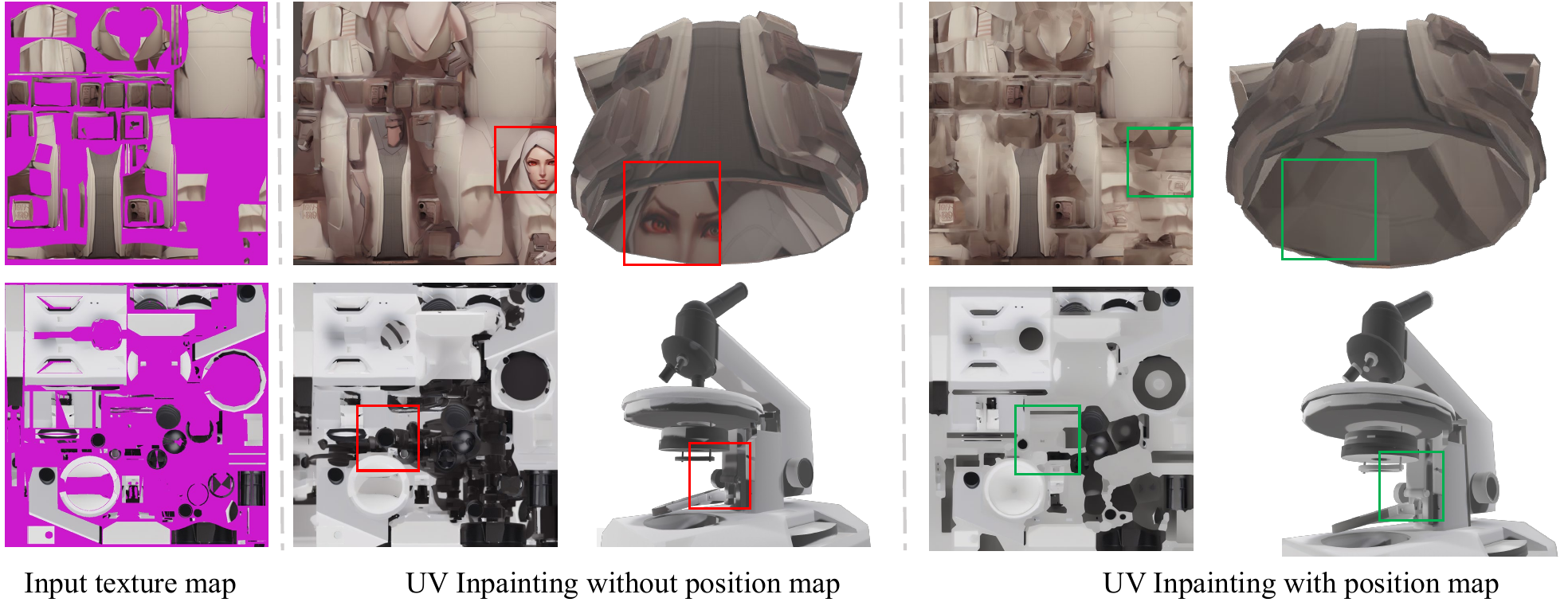}
    \caption{
    Visualization of the effect of the position map in the UV inpainting module. 
    Without the position map, the inpainted texture is semantically confused.
    The purple area indicates the uncolored area.
    }
    \label{fig:appendix:UVinpaint}
\end{figure}

\begin{figure}[h]
    \centering
    \includegraphics[width=0.95\linewidth]{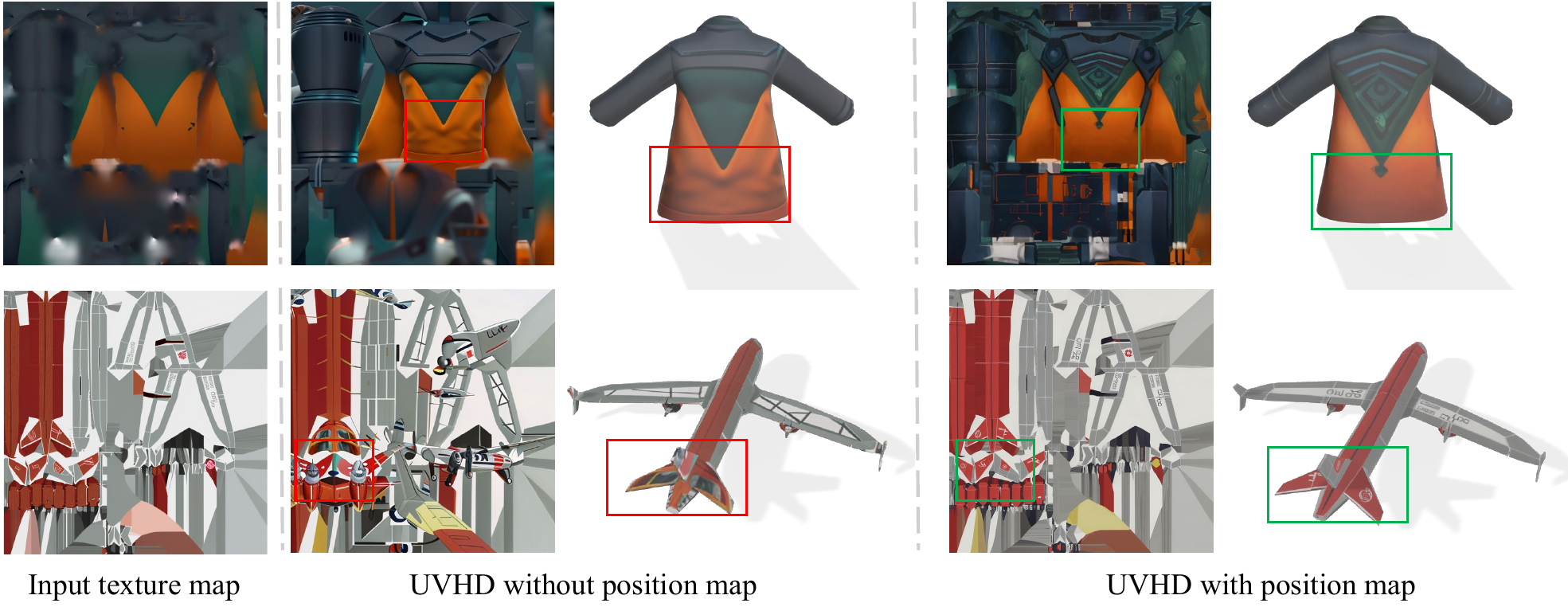}
    \caption{
    Visualization of the effect of the position map in the UVHD module. 
    In the absence of the position map, the enhanced texture appears distorted (top) or lacks semantic coherence (bottom).
    }
    \label{fig:appendix:UVHD}
\end{figure}

To demonstrate the effectiveness of position map in two texture refinement modules, UV inpainting and UVHD,
we further conduct experiments on two baselines ``UV inpainting w/o position map'' and ``UVHD w/o position map''. 
The ``UV inpainting w/o position map'' configuration refers to inpainting the uncolored area without the guidance of the position map
The ``UVHD w/o position map" configuration represents the result of enhancing the texture map in UV space, without the position map.
%
% In both scenarios, the model produces inferior results compared to our full model. 
As indicated in \cref{tab:appendix:position}, the performance shows a significant decrease when the position map is not utilized in UV inpainting or UVHD, indicating its irreplaceable function during texture refinement processing.
We visualize the results of two baselines in~\cref{fig:appendix:UVinpaint} and ~\cref{fig:appendix:UVHD}.
In both scenarios, the model produces inferior results compared to our full model. 

\begin{table}[h]
\centering
\resizebox{0.4\columnwidth}{!}{
\begin{tabular}{ccc}
\toprule
Method & {FID$\downarrow$} & {KID  $\downarrow$} \\
% & Method & {FID$\downarrow$} & {KID  $\downarrow$} \\
\hline
UV inpainting w/o position map & 39.29  & 8.36 \\
UVHD w/o position map & 37.62   & 7.96 \\
Full model & $\boldsymbol{27.28}$ & $\boldsymbol{4.81}$ \\

\bottomrule
\end{tabular}
}
\vspace{-8pt}
\caption{
Evaluation of the effectiveness of the position map in the UV Inpaint and UVHD modules.
This demonstrates the crucial role of the position map during the diffusion process in UV space.
% By integrating the generation prior in the coarse stage and the illumination-free prior in the refinement stage, our full model achieves the optimal result.
}
\vspace{-8pt}
\label{tab:appendix:position}
\end{table}

% \newpage
\subsection{Comparisons with Category-Specific Model}
\begin{figure}[h]
    \centering
    \includegraphics[width=0.95\linewidth]{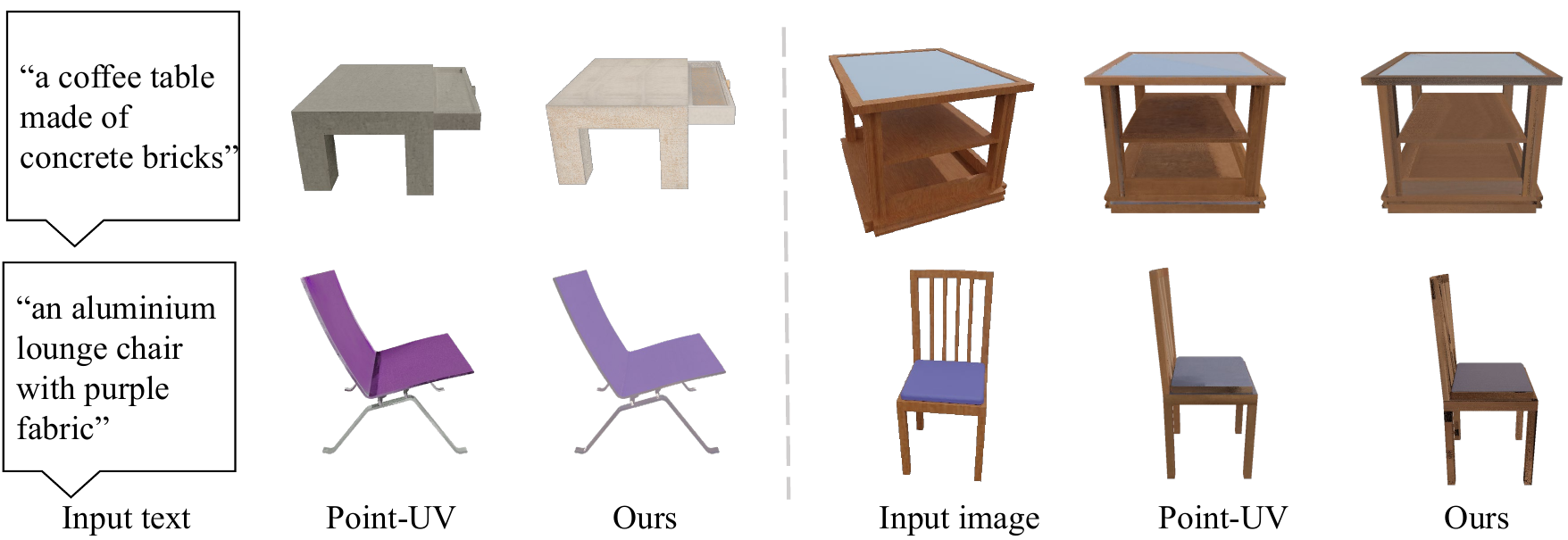}
    \caption{    
    Qualitative comparisons on texture generation conditioned under text prompt (left) and image condition (right) on ShapeNet dataset~\cite{chang2015shapenet}. 
    We compare our textured mesh against those generated by the state-of-the-art category-specific approach, Point-UV~\cite{yu2023texture}. 
    In the categories of table and chair, Paint3D achieves comparable results with Point-UV under both text and image conditions.
    }
    \label{fig:appendix:cmp}
\end{figure}

In addition, we conduct comparison experiments with a category-specific approach on the chair and table categories of ShapeNet~\cite{chang2015shapenet}.
We choose Point-UV~\cite{yu2023texture} as the baseline because 1) it represents the current state-of-the-art for category-specific texture generation, 
and 2) it has the conditional texture generation capability under both text and image conditions.
% For the image condition, we randomly render a view from the ground-truth mesh.
For the input conditions, we utilize text and images as provided in~\cite{yu2023texture}. 
As shown in~\cref{fig:appendix:cmp}, Paint3D achieves comparable results with Point-UV under both text and image conditions.

\section{Discussion on failure case}
\label{sec:appendix:failure}
Our approach still suffers from the multi-faces problem in the coarse stage which will result in a failure case. 
This issue primarily arises from the inconsistency of multi-view texture images sampled by the pre-trained 2D diffusion model, as it is not explicitly trained on multi-view datasets.
We believe that fine-tuning or retraining 2D diffusion models on large-scale multi-view datasets will improve the multi-view consistency of textures.

\begin{figure}[h]
    \centering
    \includegraphics[width=0.95\linewidth]{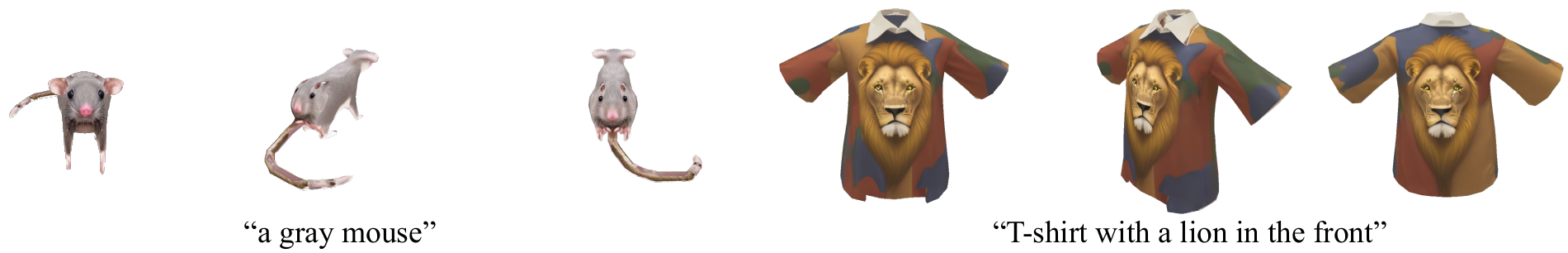}
    \caption{    
    Visualization of our failure cases. 
    Paint3D still suffers from the multi-faces problem in the coarse stage which will result in a failure case.
    Here, Paint3D generates duplicate mouse or lion faces in both the front and back views
    }
    \label{fig:appendix:failure}
\end{figure}

% \section{Inference Time}
% \label{sec:appendix:inferencetime}
% We provide a detailed ablation study on Paint3D inference time. 
% As shown in~\cref{tab:appendix:time}, the inference time of Paint3D increases linearly with the number of viewpoints.
% For texture quality, we can see that increasing the number of viewpoints can improve the quality of generated textures, but it is not that the more the viewpoints the better the results.
% % We achieve the best result when the viewpoint is set to 6.

% \begin{table}[h]
% \centering
% \resizebox{0.8\columnwidth}{!}{
% \begin{tabular}{lcccccccc}
% \toprule
% Viewpoints in Total  & \multicolumn{2}{c}{2} & \multicolumn{2}{c}{4}  & \multicolumn{2}{c}{6}  & \multicolumn{2}{c}{8}\\ \cline{2-9}
% % \multicolumn{1}{c}{Viewpoints in One Iteration} & 1 & 2 & 1 & 2 & 1 & 2 & 1 & 2 \\
% Viewpoints in One Iteration & 1 & 2 & 1 & 2 & 1 & 2 & 1 & 2 \\
% % \hline
% \toprule
% FID &  42.31 & 41.74 & 36.07 & 32.60 & 29.02 & $\boldsymbol{27.28}$ & 30.15 & 27.71 \\
% Inference Time (s) & 60.23 & 55.31 & 101.21 & 98.01 & 145.28 & 140.23 & 186.52 & 180.12 \\

% \bottomrule
% \end{tabular}
% }
% \caption{
% Evaluation of inference time costs on text-to-texture: we evaluate the total inference time to generate 1024$\times$ 1024 texture map. 
% }
% \vspace{-8pt}
% \label{tab:appendix:time}
% \end{table}

\end{document}